\documentclass{article} % For LaTeX2e
\usepackage{iclr2023_conference,times}

\usepackage{amsmath,amsfonts,bm}

\def\eqref#1{equation~\ref{#1}}
\def\1{\bm{1}}

\def\vb{{\bm{b}}}

\def\ve{{\bm{e}}}

\def\vg{{\bm{g}}}
\def\vh{{\bm{h}}}

\def\vr{{\bm{r}}}

\def\vu{{\bm{u}}}

\def\vx{{\bm{x}}}
\def\vy{{\bm{y}}}

\DeclareMathAlphabet{\mathsfit}{\encodingdefault}{\sfdefault}{m}{sl}
\SetMathAlphabet{\mathsfit}{bold}{\encodingdefault}{\sfdefault}{bx}{n}

\def\gX{{\mathcal{X}}}

\usepackage{hyperref}
\usepackage{url}
\usepackage{booktabs}
\usepackage{multirow}
\usepackage{subcaption}
\usepackage{graphicx}
\usepackage{wrapfig}
\usepackage{fancyvrb}
\usepackage{tgcursor}
\usepackage{booktabs} % for professional tables

\newcommand{\mc}{\mathcal}
\usepackage{nicefrac}
\usepackage{algorithm} 
\usepackage[ruled,vlined,linesnumbered,algo2e]{algorithm2e}
\usepackage{xcolor}
\SetKwInput{kwInit}{Init}
\SetKwInput{kwFunc}{function}
\SetKwInput{kwEnd}{end function}
\SetKwInput{kwRequire}{Require}
\SetKwComment{Comment}{$\triangleright$\ }{}

\SetCommentSty{mycommfont}

\newcommand{\STAB}[1]{\begin{tabular}{@{}c@{}}#1\end{tabular}}

\usepackage{amssymb}
\usepackage{amsthm}
\usepackage{amsmath,bm,mathtools}

\usepackage{graphicx,subcaption,ragged2e}

\title{Learning Fast and Slow for \\ Online Time Series Forecasting}

\author{Quang Pham$^1$\thanks{This work was done while the first author interned at Salesforce Research Asia.} , Chenghao Liu$^2$, Doyen Sahoo$^2$, Steven C.H. Hoi $^{1,2}$ \\
$^1$ Singapore Management University \\
\texttt{hqpham.2017@smu.edu.sg}\\
$^2$ Salesforce Research Asia\\
\texttt{\{chenghao.liu, dsahoo, shoi\}@salesforce.com }
}

\iclrfinalcopy % Uncomment for camera-ready version, but NOT for submission.
\begin{document}

\maketitle
\vspace*{-0.05in}
\begin{abstract}
Despite the recent success of deep learning for time series forecasting, these methods are not scalable for many real-world applications where data arrives sequentially. Training deep neural forecasters on the fly is notoriously challenging because of their limited ability to adapt to non-stationary environments and remember old knowledge. We argue that the fast adaptation capability of deep neural networks is critical and successful solutions require handling changes to both new and recurring patterns effectively. In this work, inspired by the Complementary Learning Systems (CLS) theory, we propose Fast and Slow learning Network (FSNet) as a novel framework to address the challenges of online forecasting. Particularly, FSNet improves the slowly-learned backbone by dynamically balancing fast adaptation to recent changes and retrieving similar old knowledge. FSNet achieves this mechanism via an interaction between two novel complementary components: (i) a per-layer adapter to support fast learning from individual layers, and (ii) an associative memory to support remembering, updating, and recalling repeating events. Extensive experiments on real and synthetic datasets validate FSNet's efficacy and robustness to both new and recurring patterns. Our code is available at \url{https://github.com/salesforce/fsnet}.
\end{abstract}

%\begin{abstract}
%The fast adaptation capability of deep neural networks is critical for online time series forecasting. Successful solutions require handling changes to both new and recurring patterns effectively. However, training deep neural forecasters on the fly is notoriously challenging because of their limited ability to adapt to non-stationary environments and remember old knowledge. In this work, inspired by the Complementary Learning Systems (CLS) theory, we propose Fast and Slow learning Network (FSNet) as a novel framework to address the challenges of online forecasting. Particularly, FSNet improves the slowly-learned backbone by dynamically balancing fast adaptation to recent changes and retrieving similar old knowledge. FSNet achieves this mechanism via an interaction between two novel complementary components: (i) a per-layer adapter to support fast learning from individual layers, and (ii) an associative memory to support remembering, updating, and recalling repeating events. Extensive experiments on real and synthetic datasets validate FSNet's efficacy and robustness to both new and recurring patterns. Our code will be made publicly available.
%\end{abstract}

\vspace*{-0.1in}
\section{Introduction}\label{sec:intro}
\vspace*{-0.05in}
Time series forecasting plays an important role in both research and industries. Correctly forecast time series can greatly benefit various business sectors such as traffic management and electricity consumption~\citep{hyndman2018forecasting}. 
As a result, tremendous efforts have been devoted to develop better forecasting models~\citep{petropoulos2020forecasting,bhatnagar2021merlion,triebe2021neuralprophet}, with a recent success of deep neural networks~\citep{li2019enhancing,xu2021autoformer,yue2021ts2vec,zhou2021informer} thanks to their impressive capabilities to discover hierarchical latent representations and complex dependencies. However, such studies focus on the batch learning setting which requires the whole training dataset to be made available a priori and implies the relationship between the input and outputs remains static throughout. This assumption is restrictive in real-world applications, where data arrives in a stream and the input-output relationship can change over time~\citep{gama2014survey}. In such cases, re-training the model from scratch could be time consuming. Therefore, it is desirable to train the deep forecaster online~\citep{anava2013online,liu2016online} using only new samples to capture the changing dynamic in the environment.

Despite the ubiquitous of online learning in many real-world applications, training deep forecasters online remains challenging for two major reasons. First, naively train deep neural networks on data streams converges slowly~\citep{sahoo2018online,aljundi2019online} because the offline training benefits such as mini-batches or training for multiple epochs are not available. Moreover, when a distribution shift happens~\citep{gama2014survey}, such cumbersome would require many more training samples to be able to learn such new concepts. Overall, deep neural networks, although possess strong representation learning capabilities, lack a mechanism to facilitate successful learning on data streams. 
Second, time series data often exhibit recurrent patterns where one pattern could become inactive and re-emerge in the future. Since deep networks suffer from the catastrophic forgetting phenomenon~\citep{mccloskey1989catastrophic}, they cannot retain prior knowledge and result in inefficient learning of recurring patterns, which further hinders the overall performance. Therefore, online time series forecasting with deep models presents a promising yet challenging problem.

To address the above limitations, we innovatively formulate online time series forecasting as an \emph{online, task-free continual learning} problem~\citep{aljundi2019online,aljundi2019task}.
Particularly, continual learning requires balancing two objectives: (i) utilizing past knowledge to facilitate fast learning of current patterns; and (ii) maintaining and updating the already acquired knowledge. These two objectives closely match the aforementioned challenges and are usually referred to as the \emph{stability-plasticity} dilemma~\citep{grossberg1982does}. 
With this connection, we develop an efficient online time series forecasting framework motivated by the \emph{Complementary Learning Systems (CLS) theory}~\citep{mcclelland1995there,kumaran2016learning}, a neuroscience framework for continual learning.
Specifically, the CLS theory suggests that humans can continually learn thanks to the interactions between the \emph{hippocampus} and the \emph{neocortex}. Moreover, the hippocampus interacts with the neocortex to consolidate, recall, and update such experiences to form a more general representation, which supports generalization to new experiences.

Motivated by the fast-and-slow learning of the CLS theory, we propose FSNet (Fast and Slow learning Network), which enhances deep networks with a complementary component to support fast learning and adaptation for online time series forecasting.
FSNet's key idea for fast learning is that it does not explicitly detect distribution shifts but instead always improving the learning of current samples. To this end, FSNet employs a per-layer adapter to model the temporal information between consecutive samples, which allow each intermediate layer to adjust itself to learn better.  In addition, FSNet further employs an associative memory~\citep{kaiser2017learning} to store important, recurring patterns observed during training. When encountering repeating events, the adapter interacts with its memory to retrieve and update the previous actions to facilitate fast learning of such patterns.
Consequently, the adapter can model the temporal smoothness in time series to facilitate learning while its interactions with the associative memory allows the model to quickly remember and continue to improve the learning of recurring patterns.
We emphasize that FSNet is a task-free method because it does not require information regarding task switches. Instead, FSNet focuses on learning the current sample by leveraging the temporal smoothness in time series and the property of recurring patterns.

In summary, our work makes the following contributions. First, we innovatively formulate learning fast in online time series forecasting with deep models as a continual learning problem. Second, motivated by the CLS theory for continual learning, we propose a fast-and-slow learning paradigm of FSNet to handle both the fast changing and long-term knowledge in time series. Lastly, we conduct extensive experiments with both real and synthetic datasets to demonstrate FSNet's efficacy and robustness.

\section{Preliminary and Related Work} \label{sec:related}
%\vspace*{-0.05in}
This section provides the necessary background of time series forecasting and continual learning.

%\vspace*{-0.05in}
\subsection{Time Series Forecasting Settings}\label{sec:ts-setting}
%\vspace*{-0.05in}
Let $\gX = (\vx_1,\ldots,\vx_T) \in \mathbb{R}^{T\times n}$ be a time series of $T$ observations, each has $n$ dimensions.
The goal of time series forecasting is that given a look-back window of length $e$, ending at time $i$: $\gX_{i,e} = (\vx_{i-e+1},\ldots,\vx_{i})$, predict the next $H$ steps of the time series as $f_{\omega}(\gX_{i,H}) = (\vx_{i+1}, \ldots, \vx_{i+H})$, where $\omega$ denotes the parameter of the forecasting model.
We refer to a pair of look-back and forecast windows as a sample.
For multiple-step forecasting ($H>1$) we follow the standard approach of employing a linear regressor to forecast all $H$ steps in the horizon simultaneously~\citep{zhou2021informer}.

\textbf{Online Time Series Forecasting }

Online time series forecasting is ubiquitous is many real-world scenarios~\citep{anava2013online,liu2016online,gultekin2018online,aydore2019dynamic} due to the sequential nature of data.
In this setting, there is no separation of training and evaluation. Instead, learning occurs over a sequence of rounds. At each round, the model receives a look-back window and predicts the forecast window. Then, the true answer is revealed to improve the model's predictions of the incoming rounds~\citep{hazan2019introduction}. The model is commonly evaluated by its accumulated errors throughout learning~\citep{sahoo2018online}.
Due to its challenging nature, online time series forecasting exhibits several challenging sub-problems, ranging from learning under concept drifts~\citep{gama2014survey}, to dealing with missing values because of the irregularly-sampled data~\citep{li2020learning,gupta2021continual}. In this work, we focus on the problem of fast learning (in terms of sample efficiency) under concept drifts by improving the deep network's architecture and recalling relevant past knowledge.

There is also a rich literature of Bayesian continual learning to address regression problems~\citep{smola2003laplace,kurle2019continual,gupta2021continual}. However, such formulation follow the Bayesian framework, which allows for forgetting of past knowledge and does not have an explicit mechanism for fast learning~\cite{huszar2017quadratic,kirkpatrick2018reply}. Moreover, such studies were not implemented with deep neural networks and it is non-trivial to extend such methods to the seting of our study. 

\subsection{Continual Learning}
\vspace*{-0.05in}
Continual learning~\citep{lopez2017gradient} is an emerging topic aiming to build intelligent agents that can learn to perform a series of tasks sequentially, with only limited access to past experiences. A continual learner must achieve a good trade-off between maintaining the acquired knowledge of previous tasks and facilitating the learning of future tasks, which is also known as the \emph{stability-plasticity} dilemma~\citep{grossberg1982does,grossberg2013adaptive}. Due to its connections to humans' learning, several neuroscience frameworks have motivated the development of various continual learning algorithms. One popular framework is the complementary learning systems theory for a dual learning system~\citep{mcclelland1995there,kumaran2016learning}.
Continual learning methods inspired from the CLS theory augments the slow, deep networks with the ability to quickly learn on data streams, either via the experience replay mechanism~\citep{lin1992self,riemer2018learning,rolnick2019experience,aljundi2019online,buzzega2020dark} or via explicit modeling of each of the fast and slow learning components~\citep{pham2021dualnet,arani2021learning}. Such methods have demonstrated promising results on controlled vision or language benchmarks. In contrast, our work addresses the online time series forecasting challenges by formulating them as a continual learning problem.
%However, real-world time series vastly differ from the controlled benchmarks because time series often exhibits strong temporal dependencies, which spans over consecutive samples. We argue that successfully capture such information is the key to facilitate learning of deep neural forecaster. As a result, we are motivated by the CLS theory to develop the novel framework of FSNet to enhance the deep neural networks with the ability to address the specific challenges of online time series forecasting.

\vspace*{-0.05in}
\section{Proposed Framework}
\vspace*{-0.05in}

\begin{figure*}[t]
	\centering
	\includegraphics[width=0.9\textwidth]{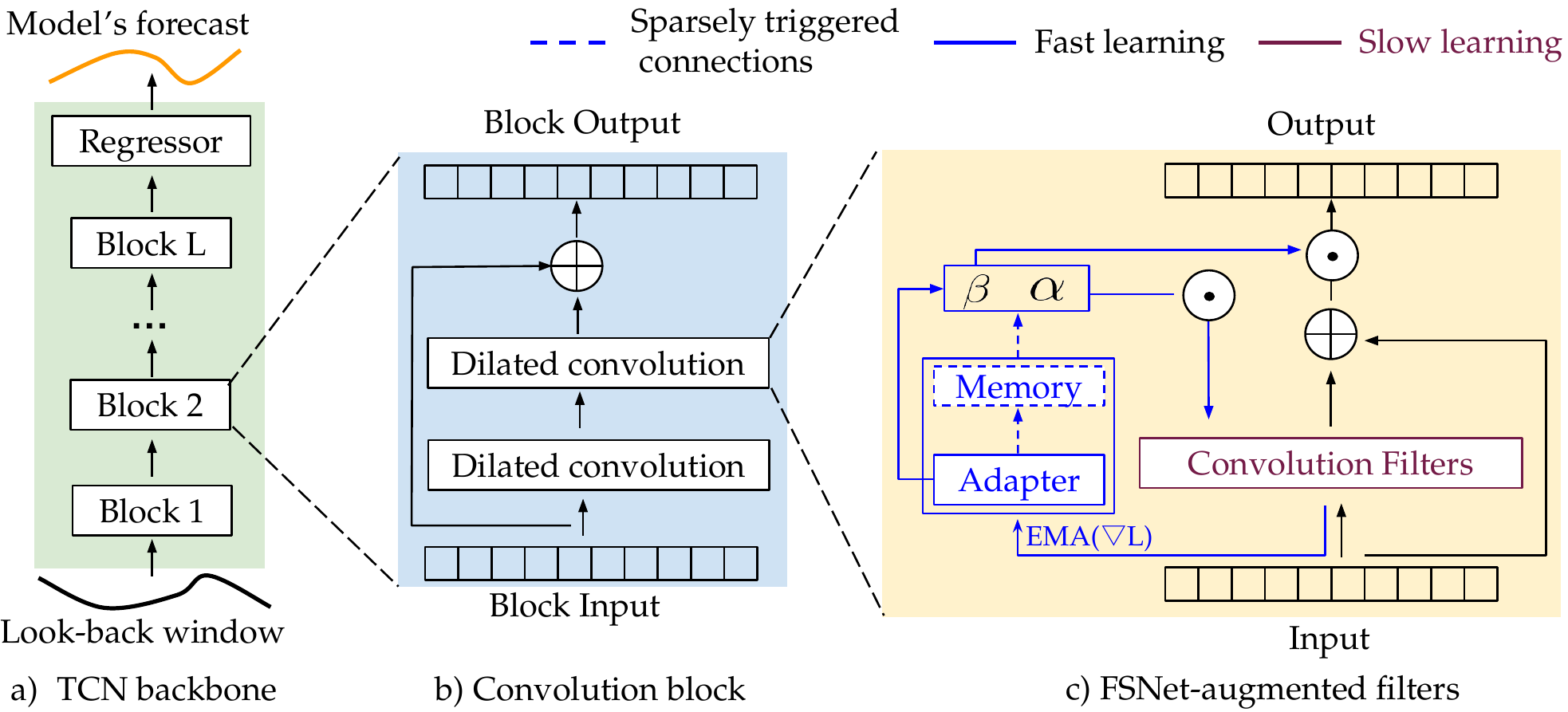}
	\caption{An overview of FSNet. a) A standard TCN backbone (green-shared) of L dilated convolution stacks (blue-shaded). b) A stack of convolution filters (yellow-shaded). c) Each convolution filter in FSNet is equipped with an adapter and associative memory to facilitate fast adaptation to both old and new patterns by monitoring the backbone's gradient EMA. Best viewed in colors.}
	\label{fig:framework}
	\vspace*{-0.1in}
\end{figure*}
This section formulates the online time series forecasting as a task-free online continual learning problem and details the proposed FSNet framework.
\subsection{Online Time Series Forecasting as a Continual Learning Problem}\label{sec:ts-cl}
Our formulation is motivated by the locally stationary stochastic processes observation, where a time series can be split into a sequence of stationary segments~\citep{vogt2012nonparametric,dahlhaus2012locally,das2016measuring}. Since the same underlying process generates samples from a stationary segment, we refer to forecasting each stationary segment as a learning task for continual learning.
We note that this formulation is general and encompasses existing learning paradigms. For example, splitting into only one segment indicates no concept drifts, and learning reduces to online learning in stationary environments~\citep{hazan2019introduction}. Online continual learning~\citep{aljundi2019online} corresponds to the case of there are at least two segments. Moreover, we also do not assume that the time points of task switch are given to the model, which is a common setting in many continual learning studies~\citep{kirkpatrick2017overcoming,lopez2017gradient}. Manually obtaining such information in real-world data can be expensive because of the missing or irregularly sampled data~\citep{li2020learning,farnoosh2021deep}.
Therefore, while we assume that the data comprises several tasks, the task-changing points are not provided, which corresponds to the online, task-free continual learning formulation~\citep{aljundi2019online,aljundi2019task,hu2020drinking,cai2021online}.

We now discuss the differences between our formulation with existing studies. First, time series evolves and old patterns may not reappear exactly in the future. Therefore, we are not interested in predicting old patterns precisely but predicting \textbf{how they will evolve}. \textit{For example, we do not need to predict the electricity consumption over the last winter. But it is more important to predict the electricity consumption this winter, assuming that it is likely to have the same pattern as the last one.}
Therefore, we do not need a separate test set for evaluation, but training follows the online learning setting where a model is evaluated by its accumulated errors throughout learning.

\vspace*{-0.1in}
\subsection{Fast and Slow Learning Networks (FSNet)}
%By reconciling online time series forecasting with continual learning, we now give a high-level overview of FSNet, which is motivated by the CLS theory.
FSNet addresses the fast adaptation to abrupt changes by the per-layer adapter (Section~\ref{sec:fast-adapter}) and facilitates learning of recurring patterns via a sparse associative memory interaction (Section~\ref{sec:ctrl-mem}).
We consider Temporal Convolutional Network (TCN)~\citep{bai2018empirical} as the backbone deep neural network to extract a time series feature representation due to the simple forward architecture and promising results~\citep{yue2021ts2vec}. The backbone has $L$ layer with parameters $\bm \theta = \{ \bm \theta_l\}_{l=1}^L$.
FSNet improves the TCN backbone with \emph{two} complementary components: a per-layer adapter $\bm \phi_l$ and a per-layer associative memory $\mc M_l$.
Thus, the total \emph{trainable} parameters is $\bm \omega = \{ \bm \theta_l, \bm \phi_l \}_{l=1}^L$ and the total associative memory is $\mc M = \{ \mc M_l\}_{l=1}^L$. Figure~\ref{fig:framework} provides an illustration of FSNet. %Although we detail FSNet with TCN, it is trivial to implement FSNet with fully connected layers, which we will experiment with in Section~\ref{sec:main-results}.
\vspace*{-0.1in}
\subsubsection{Fast-adaptation Mechanism}\label{sec:fast-adapter}
Recent works have demonstrated a \emph{shallow-to-deep} principle where shallower networks can quickly adapt to the changes in data streams or learn more efficiently with limited data~\citep{sahoo2018online,phuong2019distillation}. Therefore, it is more beneficial to learn in such scenarios with a shallow network first and then gradually increase its depth via a multi-exit architecture~\citep{phuong2019distillation}.
Our fast adaptation mechanism generalizes such approaches by allowing each layer to adapt independently rather than restricting to the network's depth. In the following, we omit the superscript indicating time indicator $t$ for brevity.

The key observation allowing for a fast, per-layer adaptation is that \emph{the partial derivative $\nabla_{\bm \theta_l}\ell$ characterizes the contribution of layer $\bm \theta_l$ to the forecasting loss $\ell$}. Since shallow networks can learn more efficiently~\citep{phuong2019distillation}, we propose to monitor and modify each layer independently to learn the current loss better.
As a result, we implement an adapter to map the layer's recent gradients to a set of smaller, more compact transformation parameters to adapt the backbone.
In online training, because of the noise and non-stationary of time series data, a gradient of a single sample can highly fluctuate~\citep{bottou1998online} and introduce noises to the adaptation coefficients. Therefore, we use the EMA of the backbone's gradient to smooth out online training's noises and to capture the temporal information in time series:
\begin{align}\label{eqn:ema}
    \hat{\vg}_{l} \gets  \gamma \hat{\vg}_{l} + (1-\gamma) \vg_l^t,
\end{align}
where $\vg_l^t$ denotes the gradient of the $l$-th layer at time $t$ and $\hat{\vg}_{l}$ denotes the EMA gradient.
The adapter takes $\hat{\vg}_{l}$ as input and maps it to the adaptation coefficients $\vu_l$.
In this work, we adopt the element-wise transformation~\citep{dumoulin2018feature-wise} as the adaptation process thanks to its simplicity and promising results in continual learning~\citep{pham2021contextual,yin2021mitigating}. Particularly, the adaptation coefficients $\vu_l$ consists of two components: a weight adaptation coefficient $\bm \alpha_l$ and (ii) a feature adaptation coefficient $\bm \beta_l$, concatenated together as $\vu_l = [\bm \alpha_l ; \bm \beta_l]$. We also absorb the bias transformation into  $\alpha_l$ for brevity. 

The adaptation process for a layer $\bm \theta_l$ involves two steps: a weight and a feature transformations, which is summarize as:
\begin{align}\label{eqn:adaptation}
    [\bm \alpha_l, \bm \beta_l] =& \vu_l,\; \mathrm{where}\; \vu_l = \bm \Omega(\hat{\vg}_l; \bm \phi_l) \\
    \tilde{\bm \theta_l} =& \mathrm{tile}(\bm \alpha_l) \odot \bm \theta_l, \; \mathrm{and} \;
    \tilde{\bm h_l} = \mathrm{tile}(\bm \beta_l) \odot \bm h_l ,\; \mathrm{where}\; \bm h_l = \tilde{\bm \theta_l} \circledast \tilde{\bm h}_{l-1}. \notag
\end{align}
Here, $\bm \theta_l$ is a stack of $I$ features maps with $C$ channels and length $Z$, $\tilde{\bm \theta_l}$ denotes the adapted weight, $\odot$ denotes the element-wise multiplication, and tile($\bm \alpha_l$) denotes that the weight adaptor is applied per-channel on all filters via a tile function that repeats a vector along the new axes. A naive implementation of Equation~\ref{eqn:adaptation} directly maps the model's gradient to the adaptation coefficients and results in a very high dimensional mapping. Therefore, we implement the \emph{chunking} operation~\citep{ha2016hypernetworks} to split the gradient into equal size chunks and then maps each chunk to an element of the adaptation coefficients. We denote this chunking operator as $\Omega(\cdot; \bm \phi_l)$ and provide the detailed description in Appendix~\ref{app:FSNet}.

\subsubsection{Remembering Recurring Events with an Associative Memory}\label{sec:ctrl-mem}
In time series, old patterns may reappear and it is imperative to leverage our past actions to improve the learning outcomes. In FSNet, an adaptation to a pattern is represented by the coefficients $\vu$, which we argue to be useful to learn repeating events. Specifically, $\vu$ represents how we adapted to a particular pattern in the past; thus, storing and retrieving the appropriate $\vu$ may facilitate learning the corresponding pattern when they reappear in the future.
Therefore, as the second key element in FSNet, we implement an associative memory to store the adaptation coefficients of repeating events encountered during learning. 
Consequently, the adapter alone can handle fast changes over a short time scale, while the associative memory can facilitate learning of repeating patterns.
In summary, we equip each adapter with an associative memory $\mc M_l \in \mathbb{R}^{N \times d}$ where $d$ denotes the dimensionality of $\vu_l$, and $N$ denotes the number of elements, which we fix as $N=32$ by default.
%The associative memory only \emph{sparsely} interacts with the adapter to store, retrieve, and update such important events, as we discuss below.

\textbf{Sparse Adapter-Memory Interactions }
Interacting with the memory at every step is expensive and susceptible to noises. Thus, we propose to trigger this interaction only when a substantial representation change happens. 
Interference between the current and past representations can be characterized in terms of a dot product between the gradients~\citep{lopez2017gradient,riemer2018learning}. 
To this end, in addition to the gradient EMA in Equation~\ref{eqn:adaptation}, we deploy another gradient EMA $\hat{\vg}'_{l}$ with a smaller coefficient $\gamma' < \gamma$ and measure their cosine similarity to trigger the memory interaction as:
\begin{equation} \label{eqn:trigger}
    \mathrm{Trigger\;if:}\; \mathrm{cos}(\hat{\vg}_l, \hat{\vg}'_l) = \frac{\hat{\vg}_l \cdot \hat{\vg}_l'}{||\hat{\vg}_l||\, ||\hat{\vg}_l||} < -\tau,
\end{equation}
where $\tau>0$ is a hyper-parameter determining the significant degree of interference.
Moreover, we want to set $\tau$ to a relatively high value (e.g. 0.7) so that the memory only remembers significant changing patterns, which could be important and may reappear. %We will discuss the hyper-parameter settings in Appendix~\ref{app:hyper}.

\textbf{The Adapter-Memory Interacting Mechanism }
Since the current adaptation coefficients may not capture the whole event, which could span over a few samples, we
perform the memory read and write operations using the adaptation coefficients's EMA (with coefficient $\gamma'$) to fully capture the current pattern. The EMA of $\vu_l$ is calculated in the same manner as Equation~\ref{eqn:ema}.
When a memory interaction is triggered, the adapter queries and retrieves the most similar transformations in the past via an attention read operation, which is a weighted sum over the memory items:
\begin{enumerate}
    \item Attention calculation: $\vr_l = \mathrm{softmax}(\mc M_l \hat{\vu}_l)$;
    \item Top-k selection: $\vr^{(k)}_l = \mathrm{TopK}(\vr_l)$;
    \item Retrieval: $\tilde{\vu_l} = \sum_{i=1}^K \vr^{(k)}_l[i]\mc M_l[i]$,
\end{enumerate}
where $\vr^{(k)}[i]$ denotes the $i$-th element of $\vr^{(k)}_l$ and $\mc M_l[i]$ denotes the $i$-th row of $\mc M_l$.
Since the memory could store conflicting patterns, we employ a sparse attention by retrieving the top-k most relevant memory items, which we fix as $k=2$. The retrieved adaptation coefficient characterizes old experiences in adapting to the current pattern in the past and can improve learning at the present time by weighted summing with the current parameters as $\vu_l \gets \tau \vu_l + (1-\tau) \tilde{\vu}_t$, where we use the same threshold value $\tau$ to determine the sparse memory interaction and the weighted sum of the adaptation coefficients. Then we perform a write operation to update and accumulate the knowledge stored in $\mc M_l$:
\begin{align}\label{eqn:mem-update}
    \mc M_l \gets& \tau \mc M_l + (1-\tau) \hat{\vu}_l \otimes \vr^{(k)}_l \; \mathrm{and} \;
    \mc M_l \gets \frac{\mc M_l}{\max(1, ||\mc M_l||_2)},
\end{align}
where $\otimes$ denotes the outer-product operator, which allows us to efficiently write the new knowledge to the most relevant locations indicated by $\vr^{(k)}_l$~\citep{rae2016scaling,kaiser2017learning}.
The memory is then normalized to avoid its values scaling exponentially. We provide FSNet's pseudo algorithm in Appendix~\ref{app:pseudo}. Lastly, we emphasize that FSNet does not reactively respond to the distribution shifts. Instead, FSNet always facilitate fast learning of deep models by improving tthe current step's learning outcome.

\vspace*{-0.1in}
\section{Experiments}
\vspace*{-0.05in}
Our experiments aim at investigating the following hypotheses: (i) FSNet facilitates faster adaptation to both new and recurring concepts compared to existing strategies with deep models; (ii) FSNet achieves faster and better convergence than other methods; and (iii) modeling the partial derivative is the key ingredients for fast adaptation. Due to space constraints, we provide the key information of the experimental setting in the main paper and provide full details, including memory analyses, additional visualizations and results in the Appendix.

\vspace*{-0.05in}
\subsection{Experimental Settings}\label{sec:exp-settings}
\vspace*{-0.05in}
\textbf{Datasets } We explore a wide range of time series forecasting datasets. {\bf ETT}\footnote{\url{https://github.com/zhouhaoyi/ETDataset}}~\citep{zhou2021informer} records the target value of ``oil temperature" and 6 power load features over a period of two years. We consider the ETTh2 and ETTm1 benchmarks where the observations are recorded hourly and in 15-minutes intervals respectively.
{\bf ECL} (Electricty Consuming Load)\footnote{\url{https://archive.ics.uci.edu/ml/datasets/ElectricityLoadDiagrams20112014}} collects the electricity consumption of 321 clients from 2012 to 2014.
{\bf Traffic}\footnote{\url{https://pems.dot.ca.gov/}} records the road occupancy rates at San Francisco Bay area freeways.
{\bf Weather}\footnote{\url{https://www.ncei.noaa.gov/data/local-climatological-data/}} records 11 climate features from nearly 1,600 locations in the U.S in an hour intervals from 2010 to 2013. 

We also construct two synthetic datasets to explicitly test the model's ability to deal with new and recurring concept drifts. We synthesize a task by sampling $1,000$ samples from a first-order autoregressive process with coefficient $\varphi$: AR$_{\varphi}$(1), where different tasks correspond to different $\varphi$ values.
The first synthetic data, {\bf S-Abrupt (S-A)}, contains abrupt, and recurrent concepts where the samples abruptly switch from one AR process to another by the following order: AR$_{0.1}$(1), AR$_{0.4}$(1), AR$_{0.6}$(1), AR$_{0.1}$(1), AR$_{0.3}$(1), AR$_{0.6}$(1). The second data, {\bf S-Gradual (S-G)} contains gradual, incremental shifts, where the shift starts at the last 20\% of each task. In this scenario, the last 20\% samples of a task is an averaged of two AR process with the order as above. Note that we randomly chose the values of $\varphi$ so that these datasets do not give unfair advantages to any methods.

\textbf{Baselines } We consider a suite of baselines from continual learning, time series forecasting, and online learning. First, the \textbf{OnlineTCN} strategy that simply trains continuously~\cite{zinkevich2003online}. Second, we consider the Experience Replay (\textbf{ER})~\cite{lin1992self,chaudhry2019tiny} strategy where a buffer is employed to store previous data and interleave old samples during the learning of newer ones. We also include three recent advanced variants of ER. First, \textbf{TFCL}~\cite{aljundi2019task} introduces a task-boundaries detection mechanism and a knowledge consolidation strategy by regularizing the networks' outputs~\cite{aljundi2018memory}. Second, \textbf{MIR}~\cite{aljundi2019online} replace the random sampling in ER by selecting samples that cause the most forgetting. Lastly, \textbf{DER++}~\cite{buzzega2020dark} augments the standard ER with a knowledge distillation strategy~\cite{hinton2015distilling}. We emphasize that ER and and its variants are strong baselines in the online setting since they enjoy the benefits of training on mini-batches, which greatly reduce noises from singe samples and offer faster, better convergence~\cite{bottou1998online}. While the aforementioned baselines use a TCN backbone, we also include \textbf{Informer}~\cite{zhou2021informer}, a recent time series forecasting method based on the transformer architecture~\cite{vaswani2017attention}. We remind the readers that online time series forecasting have not been widely studied with deep models, therefore, we include general strategies from related fields that we inspired from. Such baselines are competitive and yet general enough to extend to our problem.

\begin{table*}[t]
\setlength\tabcolsep{2.pt}
\centering
\caption{Final cumulative MSE and MAE of deep models. ``$\dagger$" indicates a transformer backbone, ``-" indicates the model did not converge. S-A: S-Abrupt, S-G: S-Gradual. Best results are in bold.}
\label{tab:main}
\small
\begin{tabular}{lccccccccccccccc}
%\begin{tabular}{@{}Sl@{}Sc@{}c@{}c@{}c@{}c@{}c@{}c@{}c@{}c@{}c@{}c@{}c@{}c@{}c@{}c}
\toprule
\multicolumn{2}{c}{Method} & \multicolumn{2}{c}{FSNet} & \multicolumn{2}{c}{DER++} & \multicolumn{2}{c}{MIR} & \multicolumn{2}{c}{ER} & \multicolumn{2}{c}{TFCL} & \multicolumn{2}{c}{OnlineTCN} & \multicolumn{2}{c}{Informer} \\  \midrule
 & H & MSE & MAE & MSE & MAE & MSE & MAE & MSE & MAE & MSE & MAE & MSE & MAE & MSE & MAE \\ \midrule
\multirow{3}{*}{\STAB{\rotatebox[origin=c]{90}{ETTh2}}} & 1 & \textbf{0.466} & \textbf{0.368} & 0.508 & 0.375 & 0.486 & 0.410 & 0.508 & 0.376 & 0.557 & 0.472 & 0.502 & 0.436 & 7.571 & 0.850 \\
 & 24 & \textbf{0.687} & \textbf{0.467} & 0.828 & 0.540 & 0.812 & 0.541 & 0.808 & 0.543 & 0.846 & 0.548 & 0.830 & 0.547 & 4.629 & 0.668 \\
 & 48 & \textbf{0.846} & \textbf{0.515} & 1.157 & 0.577 & 1.103 & 0.565 & 1.136 & 0.571 & 1.208 & 0.592 & 1.183 & 0.589 & 5.692 & 0.752 \\ \midrule
\multirow{3}{*}{\STAB{\rotatebox[origin=c]{90}{ETTm1}}} & 1 & \textbf{0.085} & \textbf{0.191} & 0.083 & 0.192 & 0.085 & 0.197 & 0.086 & 0.197 & 0.087 & 0.198 & 0.214 & 0.085 & 0.456 & 0.512 \\
 & 24 & \textbf{0.115} & \textbf{0.249} & 0.196 & 0.326 & 0.192 & 0.325 & 0.202 & 0.333 & 0.211 & {0.341} & 0.258 & 0.381 & 0.478 & 0.525 \\
 & 48 & \textbf{0.127} & \textbf{0.263} & 0.208 & 0.340 & 0.210 & 0.342 & 0.220 & 0.351 & 0.236 & 0.363 & 0.283 & 0.403 & 0.377 & 0.460 \\ \midrule
\multirow{3}{*}{\STAB{\rotatebox[origin=c]{90}{ECL}}} & 1 & 3.143 & 0.472 & \textbf{2.657} & \textbf{0.421} & 2.575 & 0.504 & 2.579 & 0.506 & 2.732 & 0.524 & 3.309 & 0.635 & - & - \\
 & 24 & \textbf{6.051} & \textbf{0.997} & 8.996 & 1.035 & 9.265 & 1.066 & 9.327 & 1.057 & 12.094 & 1.256 & 11.339 & 1.196 & - & - \\
 & 48 & \textbf{7.034} & \textbf{1.061} & 9.009 & 1.048 & 9.411 & 1.079 & 9.685 & 1.074 & 12.110 & 1.303 & 11.534 & 1.235 & - & - \\ \midrule
\multirow{2}{*}{\STAB{\rotatebox[origin=c]{90}{Traffic}}} & 1 & \textbf{0.288} & \textbf{0.253} & 0.289 & 0.248 & 0.290 & 0.251 & 0.291 & 0.252 & 0.323 & 0.273 & 0.315 & 0.283 & 0.795 & 0.507 \\
 & 24 & \textbf{0.362} & \textbf{0.288} & 0.387 & 0.295 & 0.391 & 0.302 & 0.391 & 0.302 & 0.553 & 0.383 & 0.452 & 0.363 & 1.267 & 0.750 \\ \midrule
\multirow{3}{*}{\STAB{\rotatebox[origin=c]{90}{WTH}}} & 1 & \textbf{0.162} & \textbf{0.216} & 0.174 & 0.235 & 0.179 & 0.244 & 0.180 & 0.244 & 0.177 & 0.240 & 0.206 & 0.276 & 0.426 & 0.458 \\
 & 24 & \textbf{0.188} & \textbf{0.276} & 0.287 & 0.351 & 0.291 & 0.355 & 0.293 & 0.356 & 0.301 & 0.363 & 0.308 & 0.367 & 0.370 & 0.417 \\
 & 48 & \textbf{0.223} & \textbf{0.301} & 0.294 & 0.359 & 0.297 & 0.361 & 0.297 & 0.363 & 0.323 & 0.382 & 0.302 & 0.362 & 0.367 & 0.419 \\ \midrule
\multirow{2}{*}{\STAB{\rotatebox[origin=c]{90}{S-A}}} & 1 & \textbf{1.391} & \textbf{0.929} & 2.334 & 1.181 & 2.482 & 1.213 & 2.372 & 1.157 & 2.321 & 1.144 & 2.668 & 1.216 & 3.690 & 1.410 \\
 & 24 & \textbf{1.299} & \textbf{0.904} & 3.598 & 1.439 & 3.662 & 1.450 & 3.375 & 1.360 & 3.415 & 1.366 & 3.904 & 1.491 & 3.657 & 1.426 \\ \midrule
\multirow{2}{*}{\STAB{\rotatebox[origin=c]{90}{S-G}}} & 1 & \textbf{1.760} & \textbf{1.038} & 2.335 & 1.181 & 2.482 & 1.213 & 2.476 & 1.212 & 2.428 & 1.199 & 2.927 & 1.304 & 4.024 & 1.501 \\
 & 24 & \textbf{1.299} & \textbf{0.904} & 3.598 & 1.439 & 3.662 & 1.450 & 3.667 & 1.489 & 3.829 & 1.479 & 3.904 & 1.491 & 3.657 & 1.426 \\ \bottomrule
\end{tabular}
\end{table*}

\textbf{Implementation Details } We split the data into warm-up and online training phases by the ratio of 25:75.
We follow the optimization details in~\citep{zhou2021informer} by optimizing the $\ell_2$ (MSE) loss with the AdamW optimizer~\citep{loshchilov2017decoupled}.
Both the epoch and batch size are set to {\bf one} to follow the online learning setting. 
We implement a fair comparison by making sure that all baselines use the same total memory budget as our FSNet, which includes \emph{three-times} the network sizes: one working model and two EMA of its gradient. Thus, for ER, MIR, and DER++, we allow an episodic memory to store previous samples to meet this budget. For OnlineTCN and Informer, we instead increased the backbone size.
Lastly, in the warm-up phase, we calculate the mean and standard deviation to normalize online training samples and perform hyper-parameter cross-validation.  For all benchmarks, we set the look-back window length to be $60$ and vary the forecast horizon as $H\in\{1,24,48\}$. %We 

\begin{figure*}
\captionsetup[subfigure]{justification=Centering}
\begin{subfigure}{\linewidth}
    \centering
    \includegraphics[height=0.28in,width=5.2in]{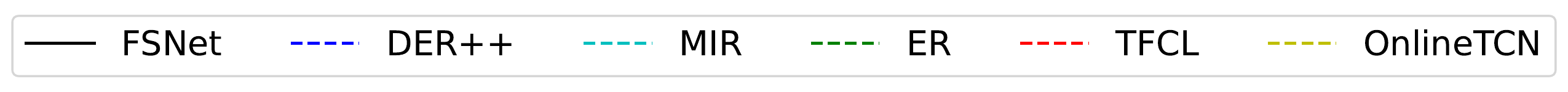}
\end{subfigure}

\subcaptionbox{ETTh2\label{fig3:etth2}}{\includegraphics[width=1.8in]{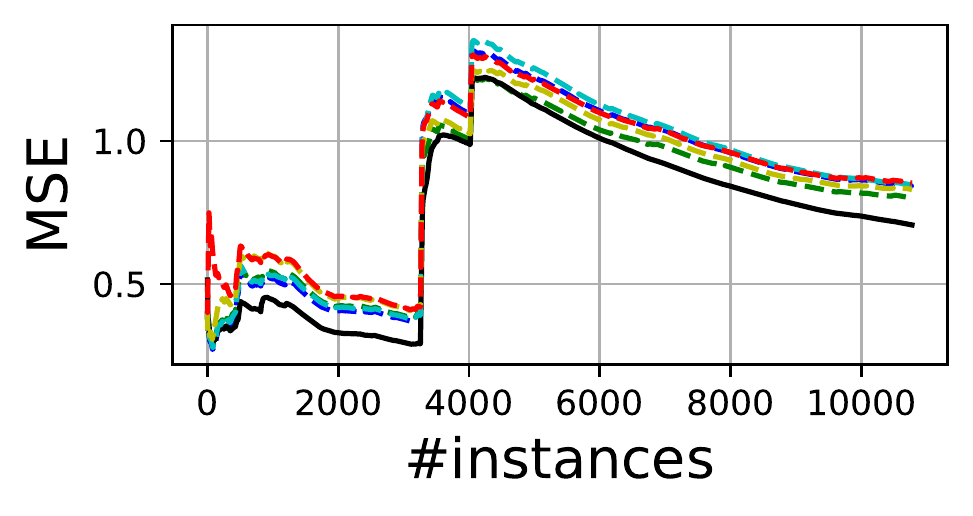}}
\subcaptionbox{ETTm1\label{fig3:ettm1}}{\includegraphics[width=1.8in]{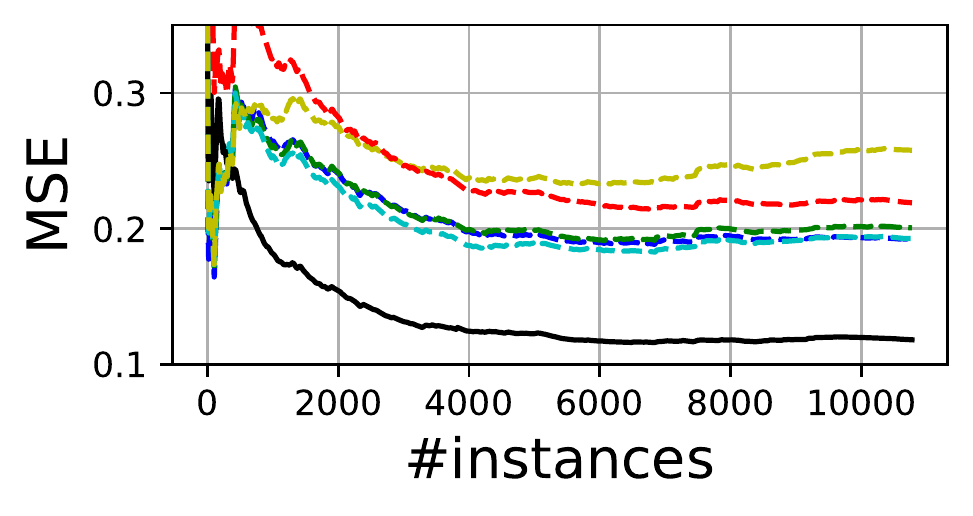}}
\subcaptionbox{ECL\label{fig3:ecl}}{\includegraphics[width=1.8in]{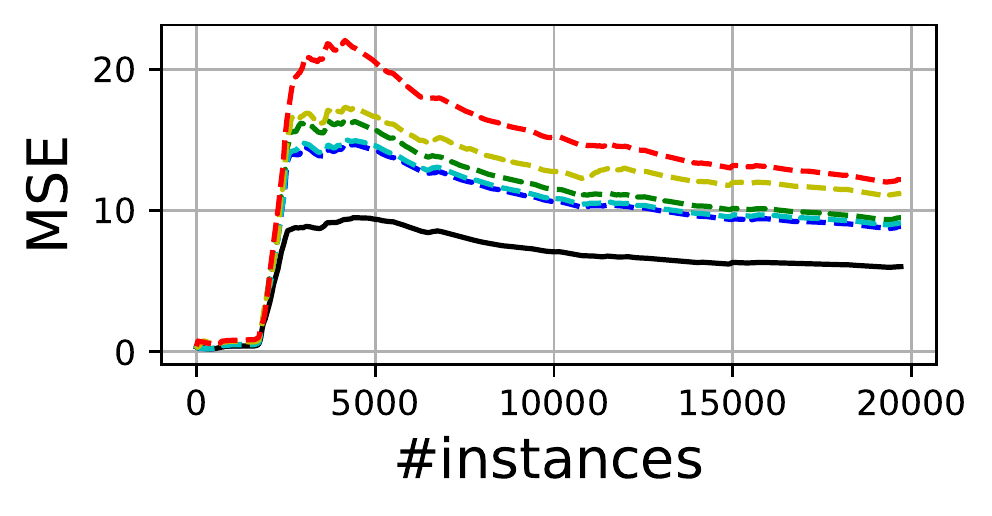}}

\subcaptionbox{Traffic\label{fig3:traffic}}{\includegraphics[width=1.8in]{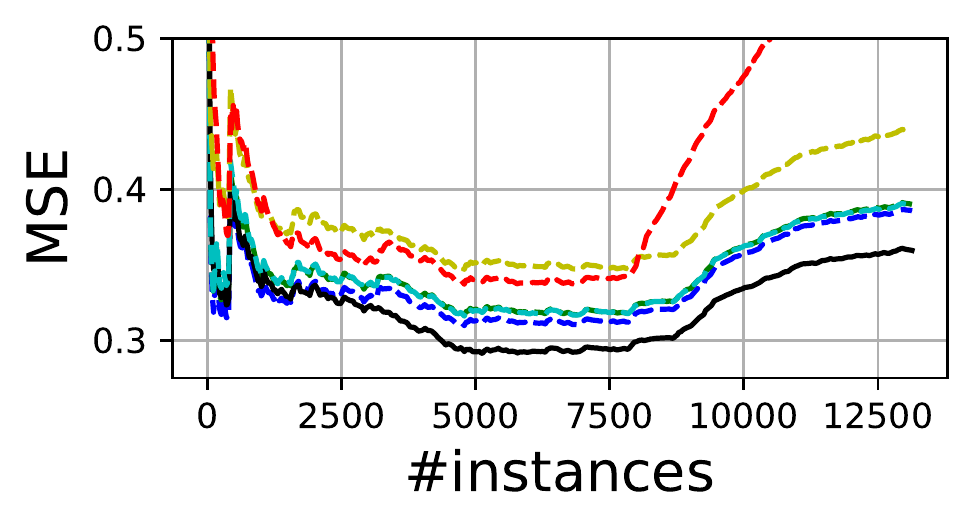}}
\subcaptionbox{WTH\label{fig3:wth}}{\includegraphics[width=1.8in]{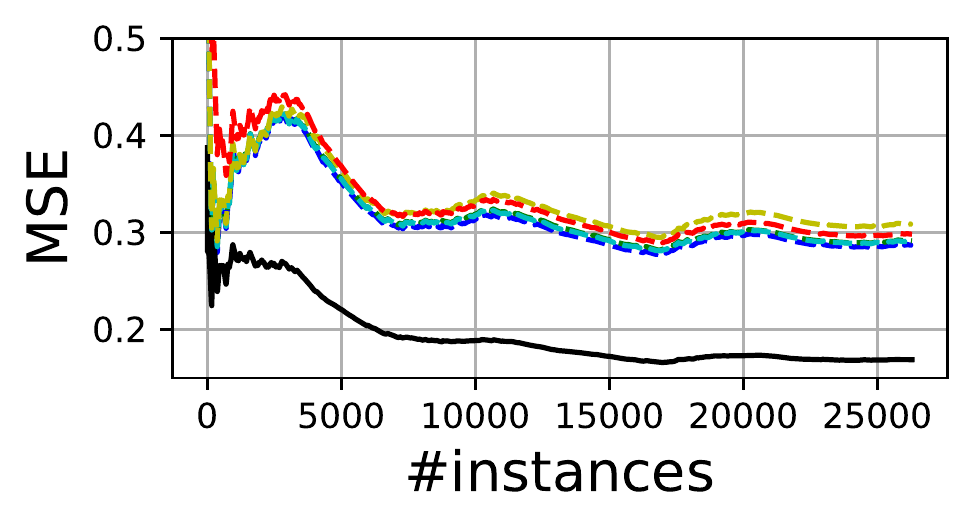}}
\subcaptionbox{S-G\label{fig3:s}}{\includegraphics[width=1.8in]{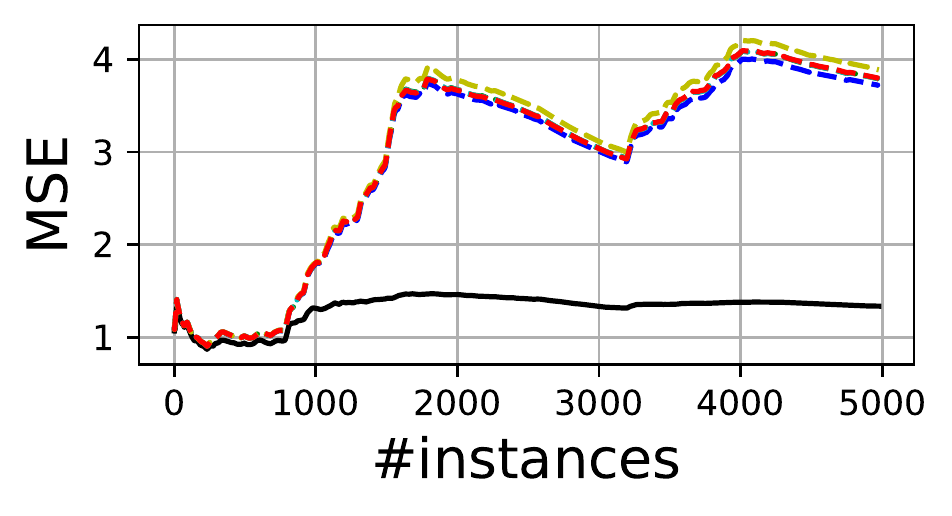}}

\caption{Evolution of the cumulative MSE loss during training with forecasting window $H=24$.}
\label{fig:curve}
\vspace*{-0.1in}
\end{figure*}

\subsection{Online Forecasting Results}\label{sec:main-results}
\textbf{Cumulative Performance }
Table~\ref{tab:main} reports the cumulative mean-squared errors (MSE) and mean-absolute errors (MAE) of deep models (TCN and Informer) at the end of training. The reported numbers are averaged over five runs, and we provide the standard deviations in the Appendix.
We observe that ER and its variants (MIR, DER++) are strong competitors and can significantly improve over the simple TCN strategies. However, such methods still cannot work well under multiple task switches (S-Abrupt). Moreover, no clear task boundaries (S-Gradual) presents an even more challenging problem and increases most models' errors. 
In addition, previous work has observed that TCN can outperform Informer in the standard time series forecasting~\citep{woo2022cost}. Here we also observe similar results that Informer does not perform well in the online setting, and is outperformed by other baselines.
On the other hand, our FSNet shows promising results on all datasets and outperforms most competing baselines across different forecasting horizons. Moreover, the significant improvements on the synthetic datasets indicate FSNet's ability to quickly adapt to the non-stationary environment and recall previous knowledge, even without clear task boundaries.

\begin{figure*}[t]
\captionsetup[subfigure]{justification=Centering}
\begin{subfigure}{\textwidth}
    \centering
    \includegraphics[height=0.3in,width=5.2in]{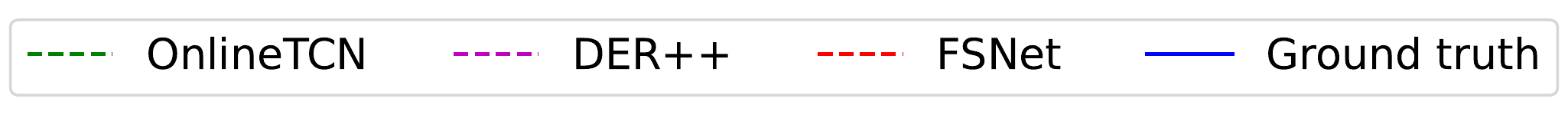}
\end{subfigure}

\subcaptionbox{Prediction from $t=3000$}{\includegraphics[width=1.8in]{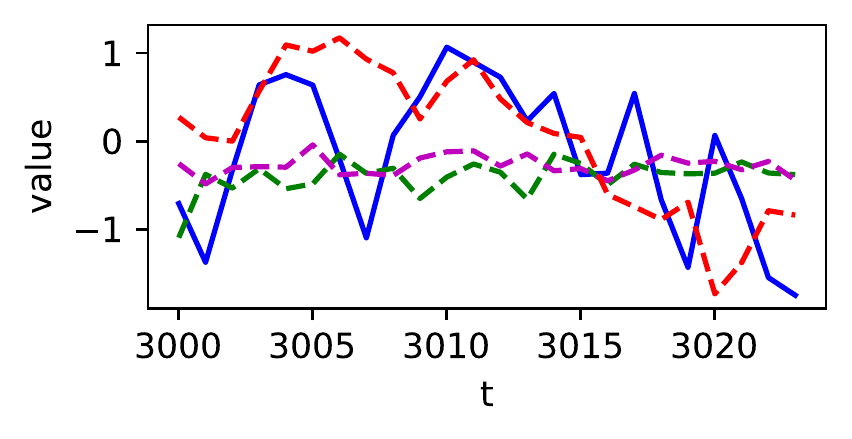}}
\subcaptionbox{Prediction from $t=3100$}{\includegraphics[width=1.8in]{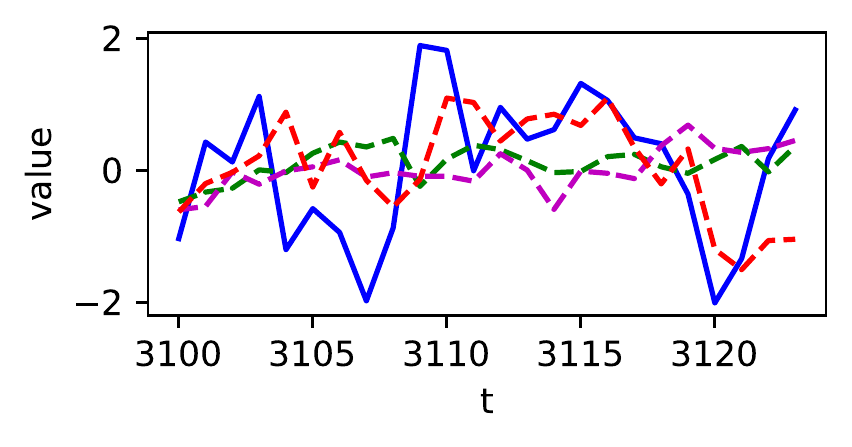}}
\subcaptionbox{Prediction from $t=3200$}{\includegraphics[width=1.8in]{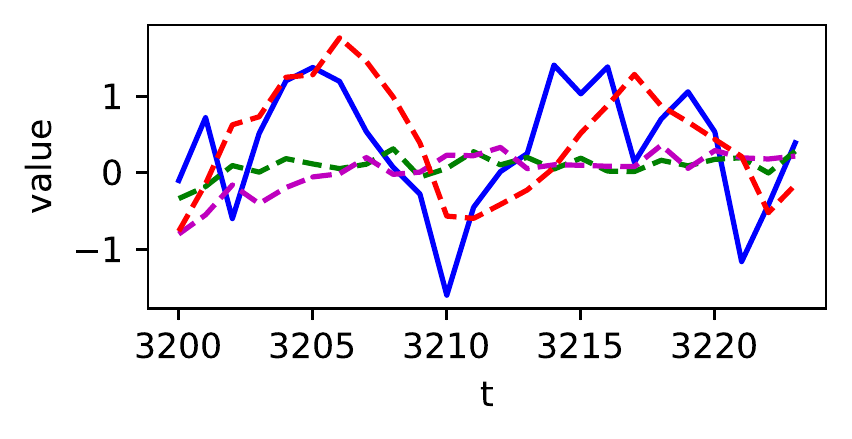}}

\subcaptionbox{Prediction from $t=4000$}{\includegraphics[width=1.8in]{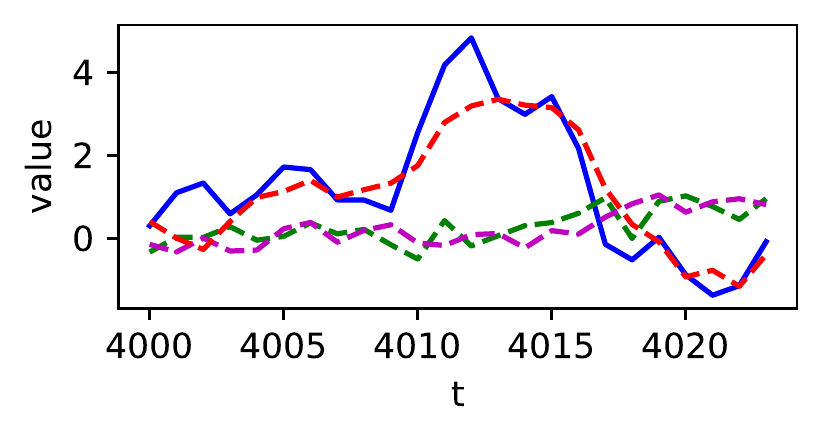}}
\subcaptionbox{Prediction from $t=4100$}{\includegraphics[width=1.8in]{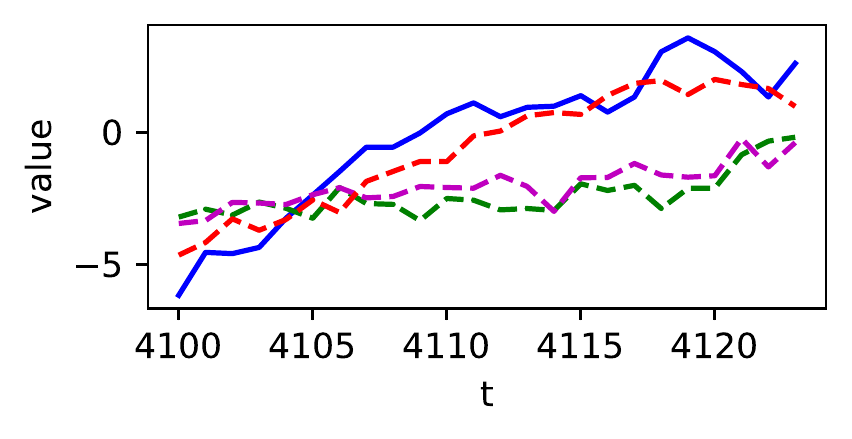}}
\subcaptionbox{Prediction from $t=4200$}{\includegraphics[width=1.8in]{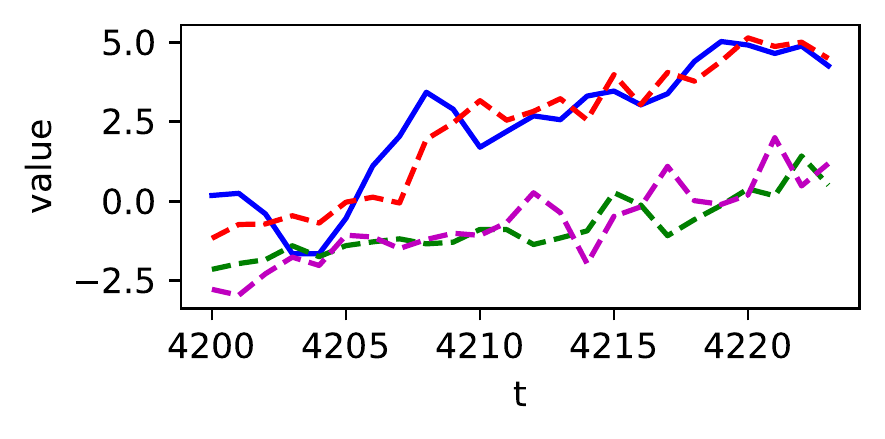}}

\subcaptionbox{Prediction from $t=5000$}{\includegraphics[width=1.8in]{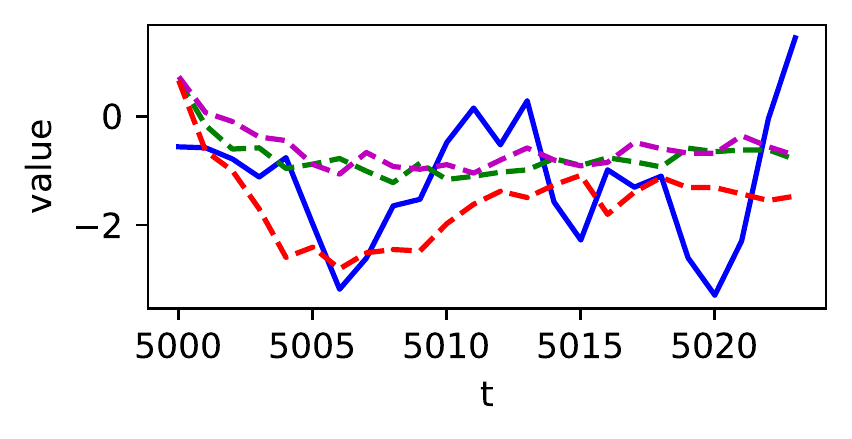}}
\subcaptionbox{Prediction from $t=5100$}{\includegraphics[width=1.8in]{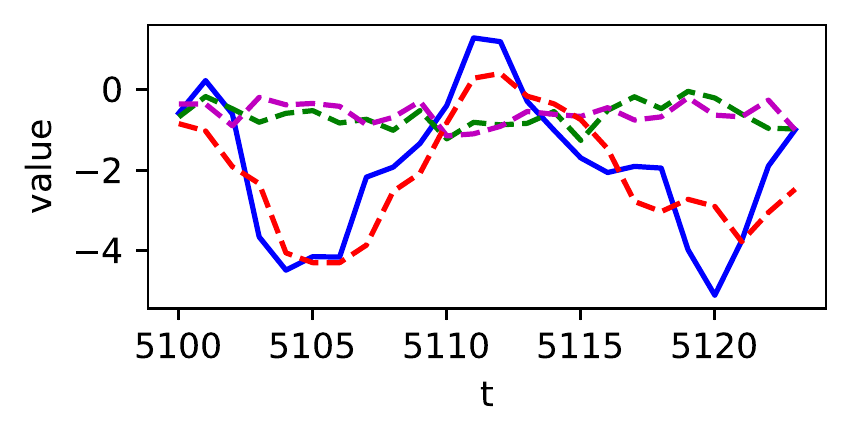}}
\subcaptionbox{Prediction from $t=5200$}{\includegraphics[width=1.8in]{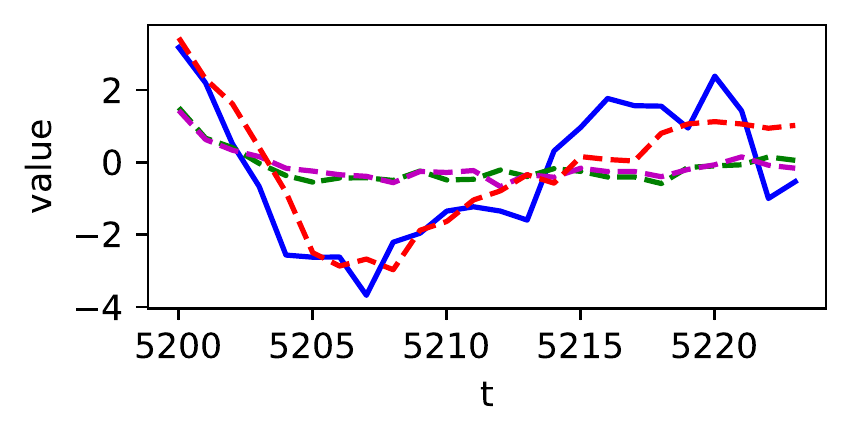}}

\caption{Visualization of the model's prediction throughout the online learning process. We focus on a short horizon of 200 time steps after a concept drift, which is critical for fast learning.}
\label{fig:quant-vis}
\vspace*{-0.25in}
\end{figure*}

\textbf{Convergent behaviors of Different Learning Strategies }
Figure~\ref{fig:curve} reports the convergent behaviors on the considered methods. We omit the S-Abrupt dataset for spaces because we observe the same behavior as S-Gradual. Interestingly, we observe that concept drifts are likely to happened in most datasets because of the loss curves' sharp peaks. Moreover, such drifts appear at the early stage of learning, mostly in the first 40\% of data, while the remaining half of data are quite stationary. This result shows that the traditional batch training is often too optimistic by only testing the model on the last data segment.
The results clearly show the benefits of ER by offering faster convergence during learning compared to OnlineTCN. However, it is important to note that storing the original data may not apply in many domains. On S-Gradual, most baselines demonstrate the inability to quickly recover from concept drifts, indicated by the increasing trend in the error curves. 
We also observe promising results of FSNet on most datasets, with significant improvements over the baselines on the ETT, WTH, and S-Gradual datasets. The remaining datasets are more challenging with missing values~\citep{li2019enhancing} and large magnitude varying \emph{within and across} dimensions, which may require calculating better data normalization statistics. While FSNet achieved encouraging results, handling the above challenges can further improve its performance. 
Overall, the results shed light on the challenges of online time series forecasting and demonstrate promising results of FSNet.

\textbf{Visualization }
We explore the model's prediction quality on the S-Abrupt since it is a univariate time series. The remaining real-world datasets are multivariate, thus challenging to visualize. Particularly, we are interested in the models' behaviour when an old task's reappear.
Therefore, in Figure~\ref{fig:quant-vis}, we plot the model's forecasting at various time points after $t=3000$. We can see the difficulties of training deep neural networks online in that the model struggles to learn at the early stages, where it only observed a few samples.  We focus on the early stages of task switches (e.g. the first 200 samples), which requires the model to quickly adapt to the distribution shifts.
With the limited samples per task and the presence of multiple concept drifts, the standard online optimization collapsed to a naive solution of predicting random noises around zero. However, FSNet can successfully capture the time series' patterns and provide better forecasts as learning progresses. Overall, we can clearly see FSNet can provide better quality forecasts compared to other baselines.

\begin{table}[t]
\small
\centering
\setlength\tabcolsep{3pt}
\caption{Final comulative MSE and MAE of different FSNet variants. Best results are in bold.}
\label{tab:ablation}
\begin{tabular}{@{\extracolsep{4pt}}lccccccccc}
\toprule
\multirow{2}{*}{Method} &    & \multicolumn{4}{c}{FSNet} & \multicolumn{4}{c}{Variant}   \\ \cmidrule{2-6} \cmidrule{7-10}
\multicolumn{2}{l}{} & \multicolumn{2}{c}{M=128}(large) & \multicolumn{2}{c}{M=32 (original)} & \multicolumn{2}{c}{No Memory} & \multicolumn{2}{c}{Naive} \\ \midrule
Data                     & H  & MSE & MAE & MSE   & MAE   & MSE   & MAE   & MSE   & MAE   \\ \midrule
\multirow{2}{*}{ETTh2}   & 24 &  {\bf 0.616}   & {\bf 0.456}    & 0.687 & 0.467 & 0.689 & 0.468 & 0.860 & 0.555 \\
                         & 48 &  {\bf 0.846}   & {\bf 0.513}    & {\bf 0.846} & 0.515 & 0.924 & 0.526 & 0.973 & 0.570 \\ \midrule
\multirow{2}{*}{Traffic} & 1  & {\bf 0.285}    &  {\bf 0.251}   & 0.288 & 0.253 & 0.294 & 0.252 & 0.330 & 0.282 \\
                         & 24 &  {\bf 0.358}   & {\bf 0.285} & 0.362 & 0.288 & 0.355 & 0.284 & 0.463 & 0.362 \\ \midrule
\multirow{2}{*}{S-A}     & 1  &  {\bf 1.388}   &  {\bf 0.928}   & 1.391 & 0.929 & 1.734 & 1.024 & 3.318 & 1.416 \\
                         & 24 &  {\bf 1.213}   & {\bf 0.870}    & 1.299 & 0.904 & 1.390 & 0.933 & 3.727 & 1.467 \\ \midrule
\multirow{2}{*}{S-G}     & 1  & {\bf 1.758}    & {1.040}    & 1.760 & {\bf 1.038} & 1.734 & 1.024 & 3.318 & 1.414 \\
                         & 24 &  {\bf 1.293}   & {\bf 0.902}    & 1.299 & 0.904 & 1.415 & 0.940 & 3.748 & 1.478 \\ \bottomrule
\end{tabular}
\vspace*{-0.1in}
\end{table}

\vspace*{-0.1in}
\subsection{Ablation Studies of FSNet's Design}\label{sec:exp:ablation}
This experiment analyzes the contribution of each FSNet's component. First, we explore the benefits of using the associative memory (Section~\ref{sec:ctrl-mem}) by constructing a \emph{No Memory} variant that only uses an adapter, without the memory. Second, we further remove the adapter, which results in the \emph{Naive} variant that directly trains the adaptation coefficients $\vu$ jointly with the backbone. The Naive variant demonstrates the benefits of monitoring the layer's gradients, our key idea for fast adaptation (Section~\ref{sec:fast-adapter}).
Lastly, we explore FSNet's scalability by increasing the associative memory size from 32 items (original) to a larger scale of 128 items.  

We report the results in Table~\ref{tab:ablation}.
We first observe that FSNet achieves similar results with the No Memory variant on the Traffic and S-Gradual datasets. One possible reason is the insignificant representation interference in the Traffic dataset and the slowly changing representations in the S-Gradual dataset. In such cases, the representation changes can be easily captured by the adapter alone and may not trigger the memory interactions.
In contrast, on ETTh2 and S-Abrupt, which may have sudden drifts, we clearly observe the benefits of storing and recalling the model's past action to facilitate learning of repeating events.
Second, the Naive variant does not achieve satisfactory results, indicating the benefits of modeling the temporal smoothness in time series via the use of gradient EMA. Lastly, the large memory variant of FSNet provides improvements in most cases, indicating FSNet's scalability with more budget.
Overall, these results demonstrated the complementary of each FSNet's components to deal with different types of concept drift in time series.

\vspace*{-0.1in}
\section{Conclusion}\label{sec:conclu}
\vspace*{-0.05in}
We have investigated the limitations of training deep neural networks for online time series forecasting in non-stationary environments, where they lack the capability to adapt to new or recurring patterns quickly. We then propose Fast and Slow learning Networks (FSNet) by extending the CLS theory for continual learning to online time series forecasting. FSNet augments a neural network backbone with two key components: (i) an adapter for adapting to the recent changes; and (ii) an associative memory to handle recurrent patterns. Moreover, the adapter sparsely interacts with its memory to store, update, and retrieve important recurring patterns to facilitate learning of such events in the future. Extensive experiments demonstrate the FSNet's capability to deal with various types of concept drifts to achieve promising results in both real-world and synthetic time series data.

\section*{Ethic Statement}
In this work, we used the publicly available datasets for experiments. We did not collect human or animal data during this study. Due to the abstract nature of this work, our method does not raise concerns regarding social/gender biases or privacy.

\section*{Reproducibility Statement}
In the supplementary materials, we included our implementations to replicate the results of our main experiment in Table~\ref{tab:main}.

\bibliography{iclr2023_conference}
\bibliographystyle{iclr2023_conference}

\appendix
\section{Appendix}
\section{Extended Related Work}\label{app:extended-rw}
This section is an extended version of Section~\ref{sec:related} where we discuss in more details the existing time series forecasting and continual learning studies.
\subsection{Time Series Forecasting}
Time series forecasting is an important problem and has been extensive studied in the literature.
Traditional methods such as ARMA, ARIMA~\citep{box2015time}, and the Holt-Winters seasonal methods~\citep{holt2004forecasting} enjoy theoretical guarantees. However, they lack the model capacity to model more complex interactions of real-world data. As a result, they cannot handle the complex interactions among different dimensions of time series, and often achieve inferior performances compared to deep neural networks on multivariate time series data~\citep{zhou2021informer,oreshkin2019n}.

Recently, learning a good time series representation has shown promising results, and deep learning models have surpassed such traditional methods on large scale benchmarks~\citep{rubanova2019latent,zhou2021informer,oreshkin2019n}. Early deep learning based approaches built upon a standard MLP models~\citep{oreshkin2019n} or recurrent networks such as LSTMs~\citep{salinas2020deepar}. Recently, temporal convolution~\citep{yue2021ts2vec} and transformer~\citep{li2019enhancing,xu2021autoformer} networks have shown promising results, achieving promising on a wide range of real-world time series. 
{\it However, such methods assume a static world and information to forecast the future is fully provided in the look-back window. As a result, they lack the ability to remember events beyond the look-back window and adapt to the changing environments on the fly.} In contrast, our FSNet framework addresses these limitation by a novel adapter and memory components.

\subsection{Continual Learning}
Human learning has inspired the design of several strategies to enable continual learning in neural networks. One successful framework is the \emph{Complementary Learning Systems} theory~\citep{mcclelland1995there,kumaran2016learning} which decomposes learning into two processes of learning fast (hippocampus) and slow (neocortex). While the hippocampus can quickly change and capture the current information, possibly with the help of experience replay, the neocortex does not change as fast and only accumulate general knowledge. The two learning systems interacts via a knowledge consolidation process, where recent experiences in the hippocampus are transferred to the neocortex to form a more general representation. In addition, the hippocampus also queries information from the neocortex to facilitate the learning of new and recurring events.
The CLS theory serves as a motivation for several designs in continual learning such as experience replay~\citep{chaudhry2019tiny}, dual learning architectures~\citep{pham2021dualnet,arani2021learning}.
In this work, we extend the fast and slow learning framework of the CLS theory to the online time series forecasting problem.

\subsection{Comparison With Existing Continual Learning for Time Series Formulations}\label{app:compare}
This section provides a more comprehensive comparison between our formulation of online time series forecasting and existing studies in~\citep{he2021clear,jaeger2017using,gupta2021continual,kurle2019continual}.

We first summarize the scope of our study. We mainly concern the online time series forecasting problem~\citep{liu2016online} and focus on addressing the challenge of fast adaptation to distribution shifts in this scenario. Particularly, when such distribution shifts happen, the model is required to take less training samples to achieve low errors, either by exploiting its representation capabilities or reusing the past knowledge. We focus on the class of deep feedforward neural network, particularly TCN, thanks to its powerful representation capabilities and ubiquitous in sequential data applications~\citep{bai2018empirical}.

CLeaR~\citep{he2021clear} also attempted to model time series forecasting as a continual learning problem. However, CLeaR focuses on accumulating knowledge over a data stream without forgetting and does not concern about a fast adaptation under distribution shifts. Particularly, CLeaR's online training involves \textbf{periodically} calibrating the pre-trained model on new out-of-distribution samples using a continual learning strategy. Moreover, CLeaR only calibrates the model when a buffer of novel samples are filled. As a result, when a distribution shifts, CLeaR could suffer from arbitrary high errors until it accumulates enough samples for calibrating. Therefore, CLeaR is not applicable to the online time series forecasting problem considered in our study. 

%Similarly, GRS~\citep{kurle2019continual} focuses on devising a framework to train Bayesian deep neural networks on data streams under the presence of concept drifts. Although achieved promising results, GRS follows the Bayesian approximation framework for continual learning~\citep{huszar2017quadratic}, which automatically determines the trade-offs between remembering old knowledge versus learning the new ones. Consequently, such a framework might not work well on more challenging benchmarks where it is important to focus on one particular aspect more closely~\citep{kirkpatrick2018reply}, which is the goal of fast adaptation in our work. Moreover, GRS was built upon the VCL framework~\citep{nguyen2017variational}, which was originally developed for multi-layered perceptrons. Therefore, it is not straightforward to extend such frameworks to the more complex architecture of TCN and compare with our method.

GR-IG~\citep{gupta2021continual} also formulates time series forecasting as a continual learning problem. However, they address the challenging of variable input dimensions through time, which could arise from the introduction of new sensors, or sensor failures. Therefore, by motivating from continual learning, GR-IG can facilitate the learning of new tasks (sensors) for better forecasting. However, GR-IG does not consider shifts in the observed distributions and focus on learning new distributions that appear over time. Consequently, GR-IG is also not a direct comparison to our method.

Lastly, we also note Conceptors~\citep{jaeger2017using} as a potential approach to address the time series forecasting problem.  Conceptors are a class of neural memory that supports storing and retrieving patterns learned by a recurrent network. In this work, we choose to use the associative memory to maintain long-term patterns, which is more common for deep feed-forward architectures used in our work. We believe that with necessary adaptation, it is possible to integrate Conceptors as the memory mechanism in FSNet, which is beyond the scope of this work.

\section{FSNet Details}\label{app:FSNet}
\subsection{Chunking Operation}
In this section, we describe the chunking adapter's chunking operation to efficiently compute the adaptation coefficients. For convenient, we denote $\mathrm{vec}(\cdot)$ as a vectorizing operation that flattens a tensor into a vector; we use $\mathrm{split}(\ve, B)$ to denote splitting a vector $\ve$ into $B$ segments, each has size $\mathrm{dim}(\ve) / B$. An adapter maps its backbone's layer EMA gradient to an adaptation coefficient $\vu \in \mathbb{R}^d$ via the chunking process as:
\begin{align*}
    \hat{\vg}_l \gets& \mathrm{vec}(\hat{\vg}_{l}) \\
    [\vb_1, \vb_2,\ldots \vb_d] \gets& \mathrm{reshape}(\hat{\vg}_l; d) \\
    [\vh_1, \vh_2, \ldots, \vh_d] \gets& [\bm W^{(1)}_{\bm \phi}\vb_1, \bm W^{(1)}_{\bm \phi}\vb_2, \ldots, \bm W^{(1)}_{\bm \phi}\vb_d]  \\
    [\vu_1, \vu_2, \ldots, \vu_d] \gets& [\bm W^{(2)}_{\bm \phi}\vh_1, \bm W^{(2)}_{\bm \phi}\vh_2, \ldots, \bm W^{(2)}_{\bm \phi}\vh_d].
\end{align*}
Where we denote $\bm W^{(1)}_{\bm \phi}$ and $\bm W^{(2)}_{\bm \phi}$ as the first and second weight matrix of the adapter. In summary, the chunking process can be summarized by the following steps: (1) flatten the gradient EMA into a vector; (2) split the gradient vector into $d$ chunks; (3) map each chunk to a hidden representation; and (4) map each hidden representation to a coordinate of the target adaptation parameter $\vu$.

\subsection{FSNet Pseudo Alorithm}\label{app:pseudo}
Algorithm~\ref{alg:fsnet} provides the psedo-code for our FSNet.

\begin{algorithm}[H]
	\DontPrintSemicolon
	\SetKwFunction{algo}{FSNet}
	\SetKwFunction{proc}{Forward}\SetKwFunction{procc}{MemoryUpdate}
	\SetKwProg{myalg}{Algorithm}{}{}
	\kwRequire {Two EMA coefficients $\gamma' < \gamma$, memory interaction threshold $\tau$}
	\kwInit{backbone $\bm \theta$, adapter  $\bm \phi$, associative memory $\mc M$, regressor $\bm R$, trigger = False}
	\For{$t \gets 1$ \textbf{to} $T$ }{
	{Receive the $t-$ look-back window $\vx_t$}
	
	{$\vh_0 = \vx_t$}
	
	\For(\tcp*[f]{Forward computation over $L$ layers}){$j \gets 1$ \textbf{to} $L$} {
	{$[\bm \alpha_l, \bm \beta_l] = \vu_l,\; \mathrm{where}\; \vu_l = \bm \Omega(\hat{\vg}_l; \bm \phi_l)$} \tcp*[f]{Initial adaptation parameter}
	
	\If{$\mathrm{trigger} == \mathrm{True}$}{%
    $\tilde{\vu_l} \gets \mathrm{Read}(\hat{\vu}_l, \mc M_l)$
    
    $\mc M_l \gets \mathrm{Write}(\mc M_l, \hat{\vu_l})$ \tcp*[f]{Memory read and write are defined in Section~\ref{sec:ctrl-mem}}
    
    $\vu_l \gets \tau \vu_l + (1-\tau) \tilde{\vu}_l$ \tcp*[f]{Weighted sum the current and past adaptation parameters}
    }
	
	{$\tilde{\bm \theta_l} = \mathrm{tile}(\bm \alpha_l) \odot \bm \theta_l$} \tcp*[f]{Weight adaptation}
	
	{$\tilde{\bm h_l} = \mathrm{tile}(\bm \beta_l) \odot \bm h_l ,\; \mathrm{where}\; \bm h_l = \tilde{\bm \theta_l} \circledast \tilde{h}_{l-1}.$} \tcp*[f]{Feature adaptation}
	
	}
	{Forecast $\hat{\vy}_t = \bm R h_T$}
	
	{Receive the ground-truth $\vy$}
	
	{Calculate the forecast loss and backpropagate}
	
	{Update the regressor $\bm R$ via SGD}
	
	\For(\tcp*[f]{Backward to update the model and EMA}){$j \gets 1$ \textbf{to} $L$} {
	{Update the EMA of $\hat{\vg_l}, \hat{\vg_l}', \vu_l$}
	
	{Update $\bm \phi_l, \bm \theta_l$ via SGD}
	
	\If{$\mathrm{cos}(\hat{\vg_l}, \hat{\vg_l}) < -\tau$}{%
    trigger $\gets$ True
    }
    
	}
	}
	\caption{Fast and Slow learning Networks (FSNet)}
	\label{alg:fsnet}
\end{algorithm}

\section{Experiment Details}
\subsection{Synthetic Data}\label{app:synthetic}
We use the following first-order auto-regressive process model $AR_{\varphi}(1)$ defined as
\begin{equation}
    X_t = \varphi X_{t-1} + \epsilon_t,
\end{equation}
where $\epsilon_t$ are random noises and $X_{t-1}$ are randomly generated. The S-Abrupt data is described by the following equation:
\begin{equation}
    X_t=
    \begin{cases}
      AR_{0.1} & \mathrm{if}\ 1 < t \leq 1000 \\
      AR_{0.4} & \mathrm{if}\ 1000 < t \leq 1999 \\
      AR_{0.6} & \mathrm{if}\ 2000 < t \leq 2999 \\
      AR_{0.1} & \mathrm{if}\ 3000 < t \leq 3999 \\
      AR_{0.4} & \mathrm{if}\ 4000 < t \leq 4999 \\
      AR_{0.6} & \mathrm{if}\ 5000 < t \leq 5999. \\
    \end{cases}
  \end{equation}

The S-Gradual data is described as
\begin{equation}
    X_t=
    \begin{cases}
      AR_{0.1} & \mathrm{if}\ 1 < t \leq 800 \\
      0.5 \times (AR_{0.1} + AR_{0.4}) & \mathrm{if}\ 800 < t \leq 1000 \\
      AR_{0.4} & \mathrm{if}\ 1000 < t \leq 1600 \\
      0.5 \times (AR_{0.4} + AR_{0.6}) & \mathrm{if}\ 1600 < t \leq 1800 \\
      AR_{0.6} & \mathrm{if}\ 1800 < t < 2400 \\
      0.5 \times (AR_{0.6} + AR_{0.1}) & \mathrm{if}\ 2400 < t \leq 2600 \\
      AR_{0.1} & \mathrm{if}\ 2600 < t \leq 3200 \\
      0.5 \times (AR_{0.1} + AR_{0.4}) & \mathrm{if}\ 3200 < t \leq 3400 \\
      AR_{0.4} &  \mathrm{if}\ 3400<  t \leq 4000 \\
      0.5 \times (AR_{0.4} + AR_{0.6} & \mathrm{if}\ 4000 < t \leq 4200 \\
      AR_{0.6} & \mathrm{if}\ 4200 < t \leq 5000 \\
    \end{cases}
  \end{equation}

\subsubsection{Baseline Details}
\paragraph{Summary} We provide a brief summary of the baselines used in your experiments
\begin{itemize}
    \item \textbf{Informer}~\citep{zhou2021informer}: a transformer-based model for time-series forecasting. 
    \item \textbf{OnlineTCN} uses a standard TCN backbone~\citep{woo2022cost} with 10 hidden layers, each of which has two stacks of residual convolution filters.
    \item \textbf{ER}~\citep{chaudhry2019tiny} augments the OnlineTCN baseline with an episodic memory to store previous samples, which are then interleaved when learning the newer ones.
    \item \textbf{MIR}~\citep{aljundi2019online} replaces the random sampling strategy in ER with its MIR sampling by selecting samples in the memory that cause the highest forgetting and perform ER on these samples.
    \item \textbf{DER++}~\citep{buzzega2020dark} augments the standard ER~\citep{chaudhry2019tiny} with a $\ell_2$ knowledge distillation loss on the previous logits.
    \item \textbf{TFCL}~\citep{aljundi2019task} is a method for online, task-free continual learing. TFCL starts with as a ER procedure and also includes a MAS-styled~\citep{aljundi2018memory} regularization that is adapted for the task-free setting.
\end{itemize}
All ER-based strategies use a reservoir sampling buffer. We also tried with a Ring buffer and did not observe any significant differences.

\paragraph{Loss function} All methods in our experiments optimize the $\ell_2$ loss function defined as follows. Let $\vx$ and $\vy \in \mathbb{R}^H$ be the look-back and ground-truth forecast windows, and $\hat{\vy}$ be the model's prediction of the true forecast windows. The $\ell_2$ loss is defined as:
\begin{equation}
    \ell(\hat{\vy}_t, \vy_t) = \ell(f_{\bm \omega}(\vx_t), \vy_t) \coloneqq \frac{1}{H} \sum_{j=1}^H ||\hat{\vy}_i - \vy_i||^2
\end{equation}

\paragraph{Experience Replay baselines} We provide the training details of the ER and DER++ baselines in this section. These baselines deploy an reservoir sampling buffer of 500 samples to store the observed samples (each sample is a pair of look-back and forecast window).

Let $\mc M$ be the episodic memory storing previous samples, $\mc B_t$ be a mini-batch of samples sampled from $\mc M$. ER minimizes the following loss function:
\begin{equation}
    \mc L^{\mathrm{ER}}_t = \ell(f_{\bm \omega}(\vx_t), \vy_t) + \lambda_{\mathrm{ER}} \sum_{(\vx,\vy)\in \mc B_t} \ell(f_{\bm \omega}(\vx), \vy),
\end{equation}
where $\ell(\cdot,\cdot)$ denotes the MSE loss and $\lambda_{\mathrm{ER}}$ is the trade-off parameter of current and past examples.
DER++ further improves ER by adding a distillation loss~\citep{hinton2015distilling}. For this purpose, DER++ also stores the model's forecast into the memory and minimizes the following loss:
\begin{equation}
    \mc L^{\mathrm{DER++}}_t = \ell(f_{\bm \omega}(\vx_t), \vy_t) + \lambda_{\mathrm{ER}} \sum_{(\vx,\vy)\in \mc B_t} \ell(f_{\bm \omega}(\vx), \vy) + \lambda_{\mathrm{DER++}} \sum_{(\vx,\hat{\vy})\in \mc B_t} \ell(f_{\bm \omega}(\vx), \hat{\vy}).
\end{equation}

\subsection{Hyper-parameters Settings}\label{app:hyper}
We cross-validate the hyper-parameters on the ETTh2 dataset and use it for the remaining ones. Particularly, we use the following configuration:
\begin{itemize}
    \item Learning rate: $3e-3$ on Traffic and ECL, $1e-3$ on the remaining datasets.
    \item Adapter's EMA coefficient $\gamma = 0.9$.
    \item Gradient EMA for triggering the memory interaction $\gamma' = 0.3$.
    \item Memory triggering threshold $\tau = 0.75$.
\end{itemize}

We found that this hyper-parameter configuration matches the motivation in the development of FSNet. In particular, the adapter's EMA coefficient $\gamma=0.9$ can capture medium-range information to facilitate the current learning. Second, the gradient EMA for triggering the memory interaction $\gamma' = 0.3$ results in the gradients accumulated in only a few recent samples.  Lastly, a relatively high memory triggering threshold $\tau=0.75$ indicates our memory-triggering condition can detect substantial representation change to store in the memory. The hyper-parameter cross-validation is performed via grid search and the grid is provided below.
\begin{itemize}
	\item Experience replay batch size (for ER and DER++): $[2,4,8]$
	\item Experience replay coefficient (for ER) $\lambda_{\mathrm{ER}}$: $[0.1, 0.2, 0.5, 0.7, 1]$
	\item DER++ coefficient (for DER++) $\lambda_{\mathrm{DER++}}$: $[0.1, 0.2, 0.5, 0.7, 1]$
	\item EMA coefficient for FSNet $\gamma$ and $\gamma'$: $[0.1, 0.2, 0.3, 0.4, 0.5, 0.6, 0.7, 0.8, 0.9]$
	\item Memory triggering threshold $\tau$ : $[0.6, 0.65, 0.7, 0.75, 0.8, 0.85, 0.9]$
	\item Number of filters per layer: 64
	\item Episodic memory size: 5000 (for ER, MIR, and DER++), 50 (for TFCL)
\end{itemize}

The remaining configurations such as data pre-processing and optimizer setting follow exactly as~\citep{zhou2021informer}.

\section{Additional Results}
\subsection{Standard Deviations}
We report the standard deviation values of the comparison experiment in Table~\ref{tab:main}, which were averaged over five runs. Overall, we observe that the standard deviation values are quite small for all experiments.

\begin{table*}[t]
\setlength\tabcolsep{2.pt}
\centering
\caption{Standard deviations of the metrics in Table~\ref{tab:main}. ``$\dagger$" indicates a transformer backbone, ``-" indicates the model did not converge. S-A: S-Abrupt, S-G: S-Gradual.}
\label{tab:std}
\small
\begin{tabular}{lccccccccccccccc}
%\begin{tabular}{@{}Sl@{}Sc@{}c@{}c@{}c@{}c@{}c@{}c@{}c@{}c@{}c@{}c@{}c@{}c@{}c@{}c}
\toprule
\multicolumn{2}{c}{Method} & \multicolumn{2}{c}{FSNet} & \multicolumn{2}{c}{DER++} & \multicolumn{2}{c}{MIR} & \multicolumn{2}{c}{ER} & \multicolumn{2}{c}{TFCL} & \multicolumn{2}{c}{OnlineTCN} & \multicolumn{2}{c}{Informer} \\  \midrule
 & H & MSE & MAE & MSE & MAE & MSE & MAE & MSE & MAE & MSE & MAE & MSE & MAE & MSE & MAE \\ \midrule
\multirow{3}{*}{\STAB{\rotatebox[origin=c]{90}{ETTh2}}} & 1 & 0.018 & 0.009 & 0.022 & 0.015 & 0.019 & 0.018 & 0.018 & 0.017 & 0.030 & 0.003 & 0.011 & 0.007 & 1.370 & 0.043  \\
 & 24 & 0.014 & 0.005 & 0.024 & 0.004 & 0.017 & 0.005 & 0.007 & 0.006 & 0.005 & 0.003 & 0.017 & 0.002 & 2.254 & 0.102  \\
 & 48 & 0.128 & 0.012 & 0.143 & 0.015 & 0.130 & 0.012 & 0.141 & 0.013 & 0.279 & 0.024 & 0.147 & 0.016 & 2.088 & 0.091  \\ \midrule
\multirow{3}{*}{\STAB{\rotatebox[origin=c]{90}{ETTm1}}} & 1 & 0.003 & 0.004 & 0.003 & 0.007 & 0.005 & 0.009 & 0.005 & 0.009 & 0.004 & 0.008 & 0.003 & 0.002 & 0.088 & 0.060  \\
 & 24 & 0.002 & 0.002 & 0.002 & 0.002 & 0.005 & 0.004 & 0.003 & 0.002 & 0.006 & 0.005 & 0.002 & 0.002 & 0.035 & 0.023  \\
 & 48 & 0.003 & 0.002 & 0.003 & 0.002 & 0.006 & 0.005 & 0.004 & 0.004 & 0.010 & 0.008 & 0.002 & 0.003 & 0.020 & 0.014  \\ \midrule
\multirow{3}{*}{\STAB{\rotatebox[origin=c]{90}{ECL}}} & 1 & 0.021 & 0.001 & 0.027 & 0.002 & 0.037 & 0.013 & 0.034 & 0.011 & 0.047 & 0.011 & 0.019 & 0.002 & - &  - \\
 & 24 & 0.096 & 0.011 & 0.072 & 0.013 & 0.261 & 0.013 & 0.236 & 0.017 & 0.338 & 0.019 & 0.077 & 0.009 & - & -  \\
 & 48 & 0.105 & 0.011 & 0.146 & 0.014 & 0.143 & 0.012 & 0.320 & 0.014 & 0.253 & 0.008 & 0.122 & 0.011 & - & - \\ \midrule
\multirow{2}{*}{\STAB{\rotatebox[origin=c]{90}{Traffic}}} & 1 & 0.001 & 0.001 & 0.001 & 0.001 & 0.001 & 0.001 & 0.001 & 0.001 & 0.004 & 0.003 & 0.001 & 0.001 & 0.009 & 0.008  \\
 & 24 & 0.002 & 0.002 & 0.002 & 0.002 & 0.002 & 0.002 & 0.002 & 0.002 & 0.004 & 0.002 & 0.002 & 0.001 & 0.015 &  0.008 \\ \midrule
\multirow{3}{*}{\STAB{\rotatebox[origin=c]{90}{WTH}}} & 1 & 0.001 & 0.001 & 0.001 & 0.002 & 0.002 & 0.002 & 0.001 & 0.002 & 0.002 & 0.002 & 0.001 & 0.001 & 0.005 & 0.005  \\
 & 24 & 0.001 & 0.001 & 0.001 & 0.001 & 0.001 & 0.001 & 0.001 & 0.001 & 0.001 & 0.001 & 0.001 & 0.001 & 0.003 & 0.003  \\
 & 48 & 0.001 & 0.001 & 0.011 & 0.007 & 0.009 & 0.005 & 0.009 & 0.005 & 0.004 & 0.006 & 0.001 & 0.001 & 0.009 & 0.008  \\ \midrule
\multirow{2}{*}{\STAB{\rotatebox[origin=c]{90}{S-A}}} & 1 & 0.112 & 0.037 & 0.171 & 0.041 & 0.176 & 0.040 & 0.159 & 0.033 & 0.283 & 0.061 & 0.009 & 0.002 & 0.149 & 0.059  \\
 & 24 & 0.199 & 0.027 & 0.202 & 0.034 & 0.192 & 0.032 & 0.022 & 0.003 & 0.011 & 0.002 & 0.174 & 0.017 & 0.322 & 0.066  \\ \midrule
\multirow{2}{*}{\STAB{\rotatebox[origin=c]{90}{S-G}}} & 1 & 0.166 & 0.039 & 0.169 & 0.041 & 0.177 & 0.040 & 0.171 & 0.039 & 0.360 & 0.083 & 0.164 & 0.039 & 0.277 & 0.099  \\
 & 24 & 0.187 & 0.033 & 0.200 & 0.033 & 0.199 & 0.035 & 0.190 & 0.032 & 0.010 & 0.002 & 0.188 & 0.033 & 0.771 & 0.099  \\ \bottomrule 
\end{tabular}
\end{table*}

\begin{table}[t]
\centering
\setlength\tabcolsep{2.5pt}
\caption{Summary of the model complexity on the ETTh2 data set with forecasta window $H=24$. We report the number of floating points incurred by the backbone and different types of memory. GI = Gradient Importance (TFCL), G-EMA = Gradient Exponential Moving Average (FSNet), AM = Associative Memory (FSNet), EM = Episodic Memory (ER). }
\label{tab:mem_complexity}
\begin{tabular}{@{\extracolsep{4pt}}lccccccc}
\toprule
\multirow{2}{*}{Method} & \multicolumn{2}{c}{Model} & \multicolumn{4}{c}{Memory} & \multirow{2}{*}{Total} \\ \cmidrule{2-3} \cmidrule{4-7}
 & Backbone & Adapter & GI & G-EMA & AM & EM &  \\ \midrule
FSNet & 1,041,288 & 733,334 & N/A & 614,400 & 1,130,496 & N/A & 3,519,518 \\
ER & 1,041,288 & N/A & N/A & N/A & N/A & 2,822,400 & 3,863,688 \\
OnlineTCN & 3,667,208 & N/A & N/A & N/A & N/A & N/A & 3,667,208 \\
TFCL & 1,041,288 & N/A & 2,082,576 & N/A & N/A & 806,400 & 3,930,264 \\ \bottomrule
\end{tabular}
\end{table} 

\subsection{Complexity Comparison}
In this Section, we analyze the memory and time complexity of FSNet.
\paragraph{Asymptotic analysis} We consider the TCN forecaster used throughout this work and analyze the \emph{model, total memory, and time} complexities of the methods considered in our work.
We let $N$ denotes the number of parameters of the the convolutional layers, $E$ denotes the length of the look-back window, and $H$ denotes the length of the forecast window. 

\begin{table}[t]
\centering
\caption{Summary of the model and total memory complexity of different methods.$N$ denotes the number parameters of the convolutional layers, $H$ and $E$ denotes the look-back and forecast windows length}
\label{tab:mem-asym}
\begin{tabular}{l|c|c|c|c|c|}
\toprule
Method            & OnlineTCN & ER & MIR & DER++ & FSNet \\ \midrule
Model Complexity  & \multicolumn{5}{c}{$\mc O(N+H)$} \\
Memory Complexity & N/A          &    \multicolumn{3}{c}{$\mc O(E+H)$}   &  $\mc O(N)$ \\ 
Total Complexity &    $\mc O(N+H)$       &   \multicolumn{3}{c}{$\mc O(N+E+H)$} &   $\mc O(N+H)$   \\ 
\bottomrule
\end{tabular}
\end{table}

\paragraph{Model and Total complexity}
We analyze the model and the total memory complexity, which arises from the model and additional memory units. 

First, the standard TCN forecaster incur a $\mc O(N+H)$ memory complexity arising from $N$ parameters of the convolutional layers, and an order of $H$ parameters from the linear regressor. 

Second, we consider the replayed-based strategies, which also incur the same $\mc O(N+H)$ model complexity as the OnlineTCN. For the total memory, they use an episodic memory to store the previous samples, which costs $\mc O(E+H)$ for both methods. Additionally, TFCL stores the importance of previous parameters while MIR makes a copy of the model for its virtual update, both of which cost $\mc O(N+H)$.
Therefore, the total memory complexity of the replay strategies (ER, DER++, MIR, and TFCL) is $\mc O(N+E+H)$.

Third, in FSNet, both the per-layer adapters and the associative memory cost similar number of parameters as the convolutional layers because they are matrices with number of channels as one dimension. Therefore, asymptotically, FSNet also incurs a model and total complexity of $\mc O(N+H)$ where the constant term is small. 

Table~\ref{tab:mem-asym} summarizes the asymptotic memory complexity discussed so far.
Table~\ref{tab:mem_complexity} shows the number of parameters used of different strategies on the ETTh2 dataset with the forecast window of $H=24$. We consider the total parameters (model and memory) of FSNet as the total budget and adjust other baselines to meet the budget. As we analyzed, for FSNet, its components, including the adapter, associative memory, and gradient EMA, require an order of parameter as the convolutional layers in the backbone network. For the OnlineTCN strategy, we increases the number of convolutional filters so that it has roughly the same total parameters as FSNet. For ER and TFCL, we change the number of samples stored in the episodic memory.

\paragraph{Time Complexity}
We report the throughput (samples/second) of different methods in Table~\ref{tab:time}. We can see that ER and DER++ have high throughput (low running time) compared to others thanks to their simplicity. As FSNet introduces additional mechanisms to allow the network to take less samples to adapt to the distribution shifts, its throughput is lower than ER and DER++. Nevertheless, FSNet is more efficient than and MIR comparable to TFCL, which are two common continual learning strategies.

\begin{table}[t]
\centering
\caption{Throughput (sample/second) of different methods in our experiments with forecast window of $H=1$.}
\label{tab:time}
\begin{tabular}{lcccccc}
\toprule
Running Time & ETTh2 & ETTm1 & WTH & ECL & Traffic & S-A \\ \midrule
ER           & 46    & 46    & 43  & 42  & 39      & 46  \\
DER++        & 45    & 45    & 43  & 42  & 38      & 46  \\
TFCL         & 29    & 28    & 27  & 27  & 26      & 27  \\
MIR          & 22    & 22    & 21  & 21  & 30      & 23  \\ \midrule
FSNet        & 28    & 28    & 28  & 27  & 27      & 29  \\ \bottomrule
\end{tabular}
\end{table}

\subsection{Robustness of Hyper-parameter Settings}
\begin{table}[t]
\centering
\caption{Results of different FSNet's hyper-parameter configurations on the ETTh2 ($H=48$) and S-A ($H=24$) benchmarks.}
\label{tab:robustness}
\begin{tabular}{lcccccc}
\toprule
\multicolumn{3}{c}{Configuration} & \multicolumn{2}{c}{ETTh2} & \multicolumn{2}{c}{S-A} \\ \cmidrule{4-7}
$\gamma$ & $\gamma'$ & $\tau$ & MSE & MAE & MSE & MAE \\ \midrule
0.9 & 0.3 & 0.75 & 0.846 & 0.515 & 1.760 & 1.038 \\
0.9 & 0.4 & 0.8 & 0.860 & 0.521 & 1.816 & 1.086 \\
0.99 & 0.4 & 0.7 & 0.847 & 0.512 & 1.791 & 1.049 \\
0.99 & 0.3 & 0.8 & 0.845 & 0.514 & 1.777 & 1.042 \\ \bottomrule
\end{tabular}
\end{table}

This experiment explores the robustness of FSNet to different hyper-parameter setting. Particularly, we focus on the configuration of \emph{three} hyper-parameters: (i) the gradient EMA $\gamma$; (ii) the short-term gradient EMA $\gamma'$; and (iii) the associative memory activation threshold $\tau$. In general, we provide two guidelines to reduce the search space of these hyper-parameters: (i) setting $\gamma$ to a high value (e.g. 0.9) and $\gamma'$ to a small value (e.g. 0.3 or 0.4); (ii) set $\tau$ to be relatively high (e.g. 0.75).
We report the results of several hyper-parameter configurations in Table~\ref{tab:robustness}.
We observe that there are not significant differences among these configurations . 
It is also worth noting that we use the same configuration for all experiments conducted in this work. Therefore, we can conclude that FSNet is robust to these configurations.

\subsection{FSNet and Experience Replay}
\begin{table}[t]
\centering
\caption{Performance of FSNet with and without experience replay.} \label{tab:fsnet+er}
\begin{tabular}{llcccc}
\toprule
\multirow{2}{*}{Data} & \multirow{2}{*}{H} & \multicolumn{2}{c}{FSNet} & \multicolumn{2}{c}{FSNet+ER} \\ \cmidrule{3-6}
                         &      & MSE   & MAE   & MSE   & MAE   \\ \midrule
\multirow{3}{*}{ETTh2}   & 1  & 0.466 & 0.368 & {\bf 0.434} & {\bf 0.361} \\
                         & 24 & 0.687 & 0.467 & {\bf 0.650}  & {\bf 0.462} \\
                         & 48 & 0.846 & 0.515 & {\bf 0.842} & {\bf 0.511} \\
\multirow{2}{*}{Traffic} & 1  & 0.321 & 0.26  & {\bf 0.243} & {\bf 0.248} \\
                         & 24 & 0.421 & 0.312 & {\bf 0.350}  & {\bf 0.275} \\ \bottomrule
\end{tabular}
\end{table}
This experiment explore the complementarity between FSNet and experience replay (ER). We hypothesize that ER is a valuable component when learning on data streams because it introduces the benefits of mini-batch training to online learning. 

We implement a variant of FSNet with an episodic memory for experience replay and report its performance in Table~\ref{tab:fsnet+er}. We can see that FSNet+ER outperforms FSNet in all cases, indicating the benefits of ER, even to FSNet. However, it is important that using ER will introduce additional memory complexity and that scales with the look-back window. Lastly, in many real-world applications, storing previous data samples might be prohibited due to privacy concerns.

\subsection{Visualizations}
\subsubsection{Visualization of the synthetic datasets}
\begin{figure*}
\captionsetup[subfigure]{justification=Centering}
\subcaptionbox{S-Abrupt}{\includegraphics[width=2.7in]{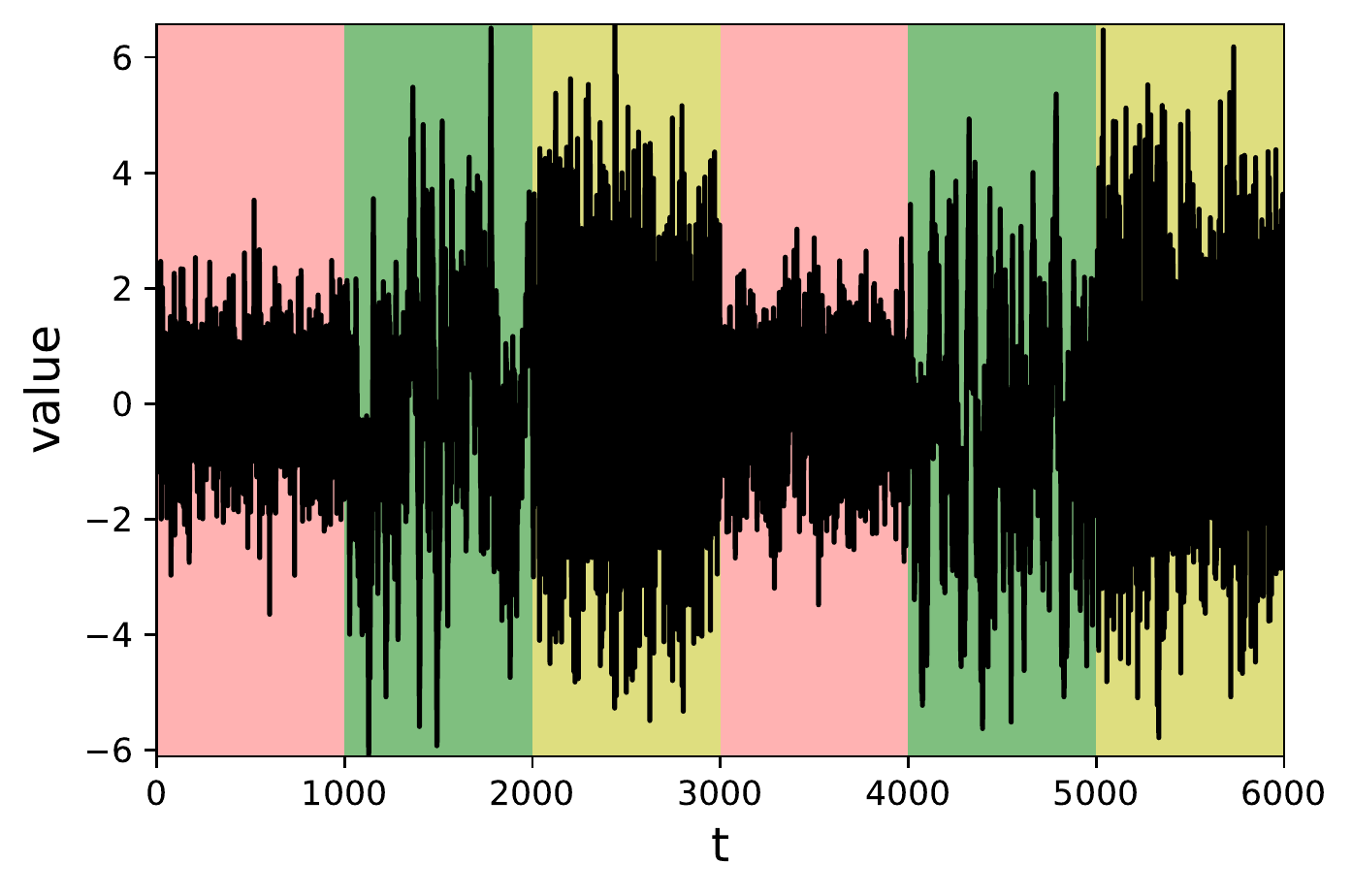}}
\subcaptionbox{S-Gradual}{\includegraphics[width=2.7in]{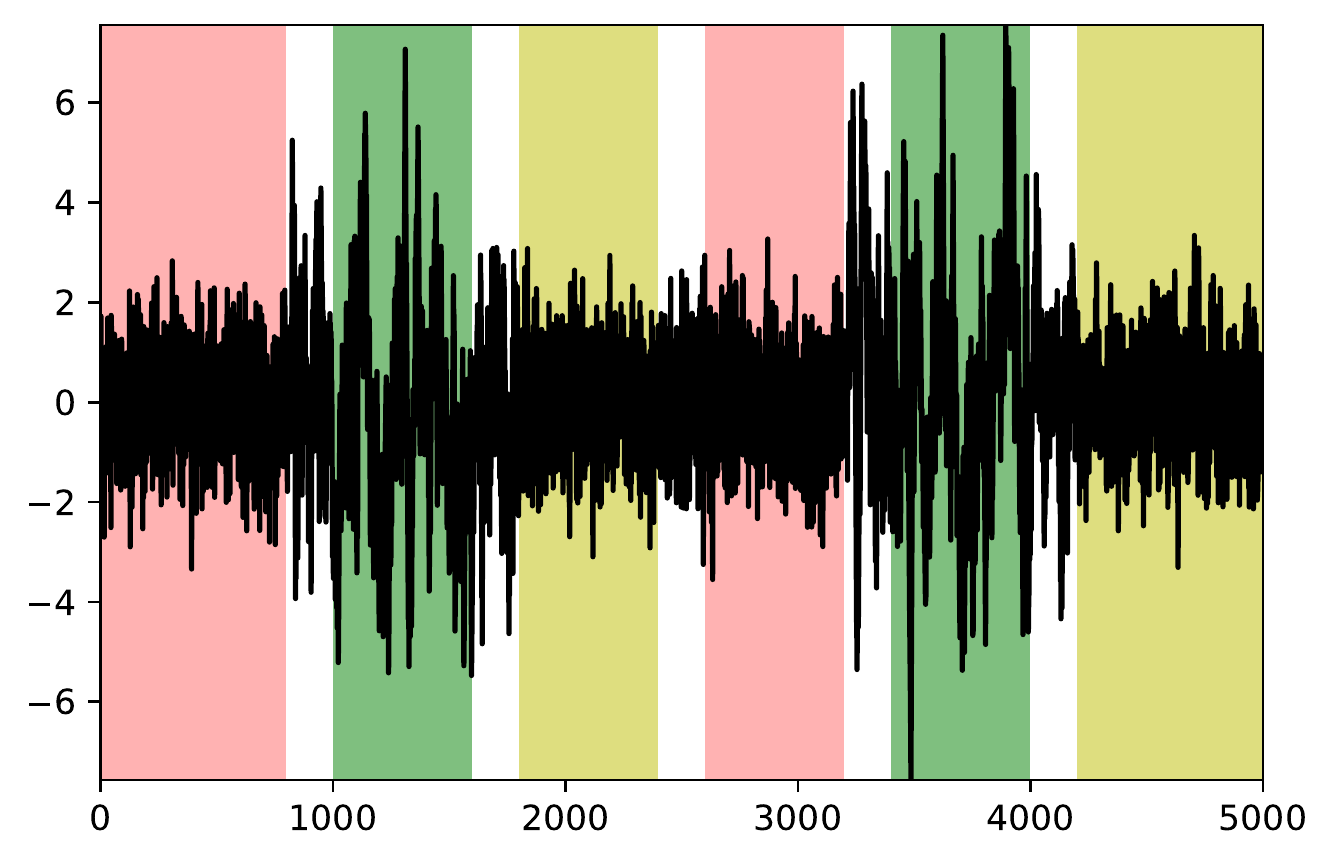}}

\caption{Visualization of the raw S-Abrupt and S-Gradual datasets before normalization. Colored regions indicate the data generating distribution where we use the same color for the same distribution. In S-Guadual, white color region indicates the gradual transition from one distribution to another. }
\label{fig:synth}
\end{figure*}

We plot the raw data (before normalization) of the S-Abrupt and S-Gradual datasets in Figure~\ref{fig:synth}.

\begin{figure*}[t]
	\centering
	\includegraphics[width=0.95\textwidth]{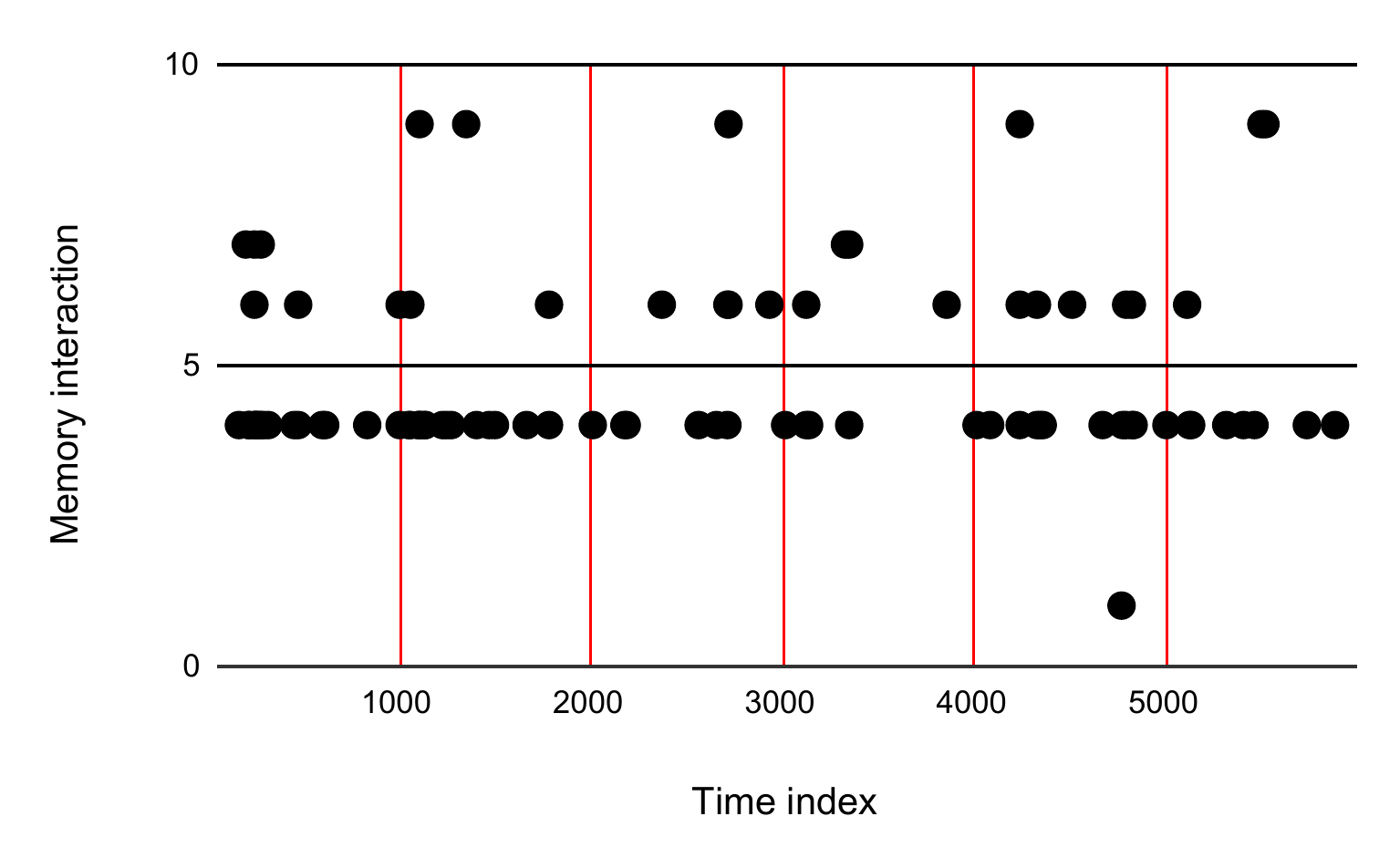}
	\caption{Activation frequency of FSNet on the S-Abrupt dataset using a TCN backbone with 20 convolutional layers. Red vertical lines indicate the time of task switches.}
	\label{fig:freq}
	\vspace*{-0.in}
\end{figure*}

\subsubsection{Activation Pattern of FSNet}
This experiment explores the associative memory activation patterns of FSNet. For this, we consider the S-Abrupt dataset and plot the activations of the associative memory units in Figure~\ref{fig:freq}. Note that the TCN backbone has 20 convolutional layers.
We remind that in S-Abrupt, the first 3,000 samples belong to three different data distribution and these distribution sequentially reappear in the last 3,000 samples. 
First, we observe that not all layers are equally important for the tasks. Particularly, FSNet mostly uses the fourth and sixth layers, and rarely uses the deeper ones.
Second, we note that the episodic memories regularly activates throughout learning to support the current time's learning outcomes. This pattern is different from change point detection models where it passively reacts to the environment, i.e., such models do not support fast learning when there is no concept/distribution shifts.

\section{Discussion and Future Work}
We discuss two scenarios where FSNet may not work well. First, we suspect that FSNet may struggle when concept drifts do not happen uniformly on all dimensions. This problem arises from the irregularly sampled time series, where each dimension is sampled at a different rate. In this scenario, a concept drift in one dimension may trigger FSNet’s memory interaction and affect the learning of the remaining ones. Moreover, if a dimension is sampled too sparsely, it might be helpful to leverage the relationship along both the time and spatial dimension for a better result.

Second, applications such as finance, which involve many complex repeating patterns, can be challenging for FSNet. In such cases, the number of repeating patterns may exceed the memory capacity of FSNet, causing catastrophic forgetting. In addition, forecasting complex time series requires the network to learn a good representation, which may not be achieved by increasing the model complexity alone. In such cases, incorporating a representation learning component might be helpful.

We now discuss several aspects for further studies. We follow Informer to apply the z-normalization per feature, which is a common strategy. This strategy works well in the batch setting because its statistics were estimated using 80\% of training data. However, after a concept drift in online learning, it is unreliable to use previous statistics (estimated over 25\% samples) to normalize samples from a new distribution. In such cases, it could be helpful to adaptively normalize samples from new distributions (using the new distribution’s statistics). This could be achieved via an online update of the normalization statistics or using a sliding window technique. In addition, while FSNet presents a general framework to forecast time series online, adopting it to a particular application requires incorporating specific domain knowledge to ensure satisfactory performances. In summary, we firmly believe that FSNet is an encouraging first step towards a general solutions for an important, yet challenging problem of online time series forecasting.

\end{document}